\documentclass[default,iicol]{sn-jnl}

\jyear{2023}%

\theoremstyle{thmstyleone}%
\theoremstyle{thmstyletwo}%

\theoremstyle{thmstylethree}%

\usepackage{graphicx}%
\usepackage{multirow}%
\usepackage{amsmath,amssymb,amsfonts}%
\usepackage{amsthm}%
\usepackage{mathrsfs}%
\usepackage[title]{appendix}
\usepackage{textcomp}%
\usepackage{manyfoot}%
\usepackage{booktabs}%
\usepackage{algorithm}%
\usepackage{algorithmicx}%
\usepackage{algpseudocode}%
\usepackage{listings}%
\usepackage{hyperref} 
\usepackage{url}
\usepackage{xurl}
\usepackage{color,colortbl}
\definecolor{Gray}{gray}{0.9}
\definecolor{top1}{RGB}{255,179,179}
\definecolor{top2}{RGB}{255,217,179}
\definecolor{top3}{RGB}{255,255,179}

\begin{document}

\title[Article Title]{RDG-GS: Relative Depth Guidance with Gaussian Splatting for Real-time Sparse-View 3D Rendering}


\author[1,3]{\fnm{Chenlu} \sur{Zhan}}\email{chenlu.22@intl.zju.edu.cn}

\author[2,3]{\fnm{Yufei} \sur{Zhang}}\email{yufei1.23@intl.zju.edu.cn}

\author[3]{\fnm{Yu} \sur{Lin}}\email{yulin@intl.zju.edu.cn}

\author*[3]{\fnm{Gaoang} \sur{Wang}}\email{gaoangwang@intl.zju.edu.cn}

\author*[3]{\fnm{Hongwei} \sur{Wang}}\email{hongweiwang@intl.zju.edu.cn}

\affil[1]{\orgdiv{College of Computer Science and Technology}, \orgname{Zhejiang University}, \orgaddress{\city{Hangzhou}, \country{China}}}

\affil[2]{\orgdiv{College of Biomedical Engineering and Instrument Science}, \orgname{Zhejiang University}, \orgaddress{\city{Hangzhou}, \country{China}}}

\affil*[3]{\orgdiv{ZJU-UIUC Institute}, \orgname{Zhejiang University}, \orgaddress{\city{Hangzhou}, \country{China}}}


\abstract{ Efficiently synthesizing novel views from sparse inputs while maintaining accuracy remains a critical challenge in 3D reconstruction. While advanced techniques like radiance fields and 3D Gaussian Splatting achieve rendering quality and impressive efficiency with dense view inputs, they suffer from significant geometric reconstruction errors when applied to sparse input views.
Moreover, although recent methods leveraging monocular depth estimation to enhance geometric learning, their dependence on single-view estimated depth often leads to view inconsistency issues across different viewpoints. Consequently, this reliance on absolute depth can introduce inaccuracies in geometric information, ultimately compromising the quality of scene reconstruction with Gaussian splats.
In this paper, we present \textbf{RDG-GS}, a novel sparse-view 3D rendering framework with \textbf{R}elative \textbf{D}epth \textbf{G}uidance based on 3D \textbf{G}aussian \textbf{S}platting.
The core innovation lies in utilizing relative depth guidance to refine the Gaussian field, steering it towards view-consistent spatial geometric representations, thereby enabling the reconstruction of accurate geometric structures and capturing intricate textures.
First, we devise refined depth priors to rectify
the coarse estimated depth and insert global and fine-grained scene information to regular Gaussians. Building on this, to address spatial geometric inaccuracies from absolute depth, we propose relative depth guidance by optimizing the similarity between spatially correlated patches of depth and images.
Additionally, we also directly deal with the sparse areas challenging to converge by the adaptive sampling for quick densification. 
 Across extensive experiments on Mip-NeRF360, LLFF, DTU, and Blender, RDG-GS demonstrates state-of-the-art rendering quality and efficiency, making a significant advancement for real-world application. }




\maketitle

\begin{figure*}[htbp]
\centering
\includegraphics[width=1\linewidth]{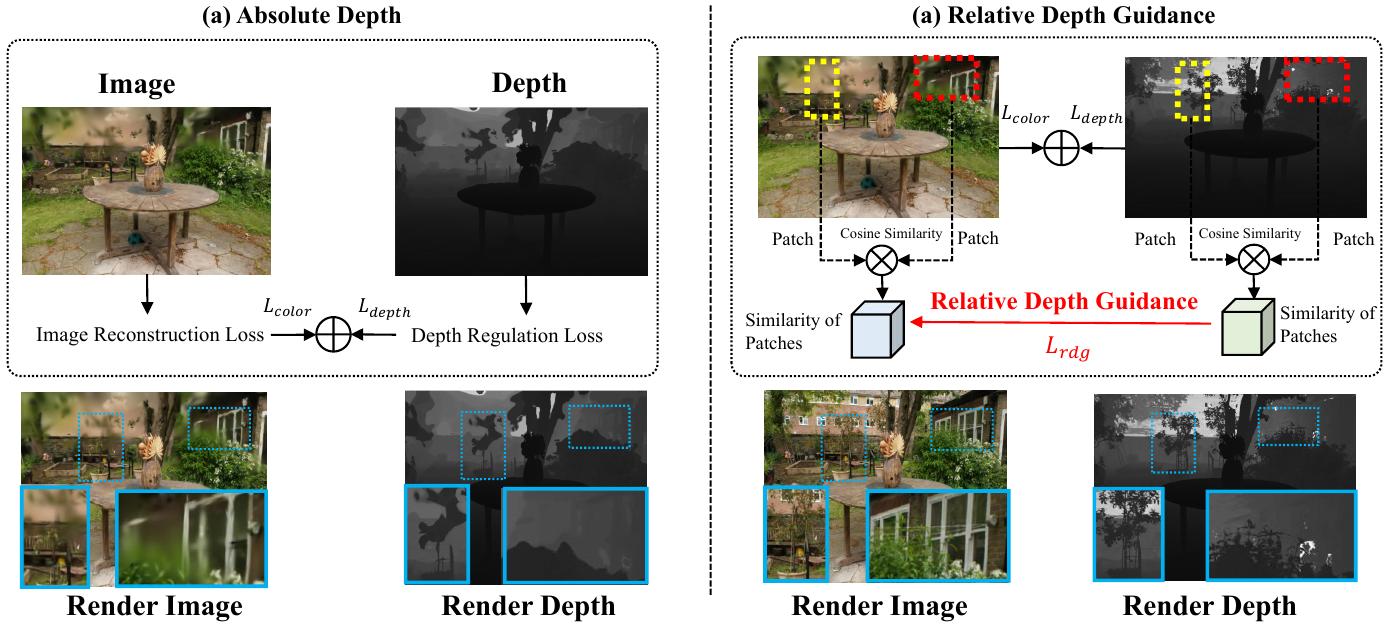}
 \small
 \caption{
(a) General Absolute Depth Method. Most methods~\cite{li2024dngaussian, zhu2023fsgs, guo2024depth} rely on monocular estimated depth, combining depth regularization and image reconstruction losses to optimize the Gaussian field. However, this approach rely on single-view depth which introduces inconsistency problems and results in erroneous geometric information, resulting in inaccurate geometric scene structures (highlighted in \textcolor{blue}{blue} boxes).
(b) Our Proposed Relative Depth Guidance: By utilizing relatively refined depth with view-consistent spatial geometric information, we compute patch-wise similarity to extract relative geometric cues for solving inconsistency, enabling accurate scene geometry reconstruction and high-quality rendering (highlighted in \textcolor{blue}{blue} boxes).}
\label{compare_gloabl}
\end{figure*}

\section{Introduction}\label{sec1}
Synthesizing novel views from sparse views is crucial for virtual reality applications (e.g. virtual

\noindent{reality and autonomous driving)~\cite{gao2022nerf,rabby2023beyondpixels,tewari2022advances}. 
Although Neural Radiance Field (NeRF)~\cite{mildenhall2021nerf} and 3D Gaussian Splatting (3D-GS)~\cite{kerbl20233dggs} are highly effective at reconstructing realistic and accurate geometric scenes under 
dense view conditions, both face significant challenges when dealing with sparse input views.} 
NeRF's capability to recover fine geometric details is hindered by its reliance on extensive view coverage~\cite{wang2023sparsenerf}, and its computationally intensive training and rendering processes limit practicality~\cite{muller2022instant,fridovich2022plenoxels,wang2022fourier}. Similarly, 3D-GS, renowned for achieving real-time rendering capabilities via efficient 3D differentiable splatting, has demonstrated remarkable advancements in rendering speed and computational efficiency. However, its performance remains highly contingent upon the quantity and quality of the initially sampled Gaussian primitives.  Sparse input views exacerbate issues like geometric degradation and over-smoothing, compromising its ability to reconstruct fine structures and maintain scene fidelity accurately.


To address this, some 3D-GS based works~\cite{zhu2023fsgs,chung2023depth,li2024dngaussian}, inspired by NeRF-based approaches~\cite{wang2023sparsenerf,yang2023freenerf,yu2021pixelnerf} designed for sparse-view setting, leverage coarse depth priors from a monocular depth estimator to enforce depth regularization. By introducing this form of supervision, these approaches attempt to mitigate the inherent ambiguities in sparse-view reconstruction and enhance the reliability of the depth representation.
However, despite the improvements brought about by leveraging monocular depth estimated priors, there still exist some \textbf{non-trivial challenges}, as outlined below:
(1) \textbf{Coarse estimated depth.} Existing works~\cite{li2024dngaussian,zhu2023fsgs,guo2024depth} directly utilize estimated depth priors generated by monocular depth estimators as
supervision, but ignoring the estimated depth exists unavoidable estimation errors and inherent ambiguity, especially for the global structure and boundary areas of the scene. Applying coarse depth may mislead the scene into erroneous and oversmooth shapes, thus damaging the reconstruction of splats. 
(2) \textbf{Single-view  inconsistent depth.} 
While existing methods~\cite{wang2023sparsenerf,zhu2023fsgs,li2024dngaussian} leverage single-view estimated depth to guide Gaussian geometry optimization, these depths rely solely on absolute geometric information, often introducing the view inconsistency that compromises the reconstruction quality of Gaussian splats. Besides, 3D-GS~\cite{kerbl20233dggs} focuses on optimizing absolute Gaussian splats throughout scenes, but lacks spatially relative geometric information, which may hinder the model’s ability to capture global geometric relationships and cause it to fall into local optima.
(3) \textbf{Inadequate and sparse initialization of 3D-GS.} Under sparse-view settings, inadequate and coarse initialization of 3D-GS results in a sharp decline of details and blurred geometry in the far and boundary areas while slowing down the rendering speed.

In this paper, we propose the \textbf{RDG-GS},  a novel sparse-view approach that leverages \textbf{R}elative \textbf{D}epth-\textbf{G}uidance based on \textbf{G}aussian \textbf{S}platting to enable high-quality 3D reconstruction and real-time rendering. The key innovation lies in leveraging the relative depth guidance to refine the Gaussian field, directing it toward view-consistent spatial geometry, and enabling accurate geometric reconstructions while capturing intricate textures.
We first propose refined depth priors to address estimation errors, integrating global and fine-grained contexts of high-quality images.
Notably, we propose relative depth guidance to provide view-consistent spatial relative geometry. This method optimizes the similarity between spatially correlated patches of depth and images, as illustrated in Fig.~\ref{compare_gloabl}.
Besides, under sparse-view inputs, enhancing both the quality and quantity of initialization points for 3D-GS becomes essential. To address this, we introduce an adaptive sampling strategy that significantly enhances densification.
Experiments in scene-level and object-level datasets validate the effectiveness and efficiency of RDG-GS in real-world applications with superior reconstruction quality and real-time rendering speed.
Our contributions are summarized as follows:
\begin{itemize}
\item We present RDG-GS, a novel sparse-view 3D reconstruction model that utilizes relative depth guidance by optimizing the spatial depth-image similarity,  thereby ensuring view-consistent geometry reconstruction and fine-grained refinement.

\item We integrate the global and local scene information into Gaussians through refined depth prior for accurate geometry and fine-grained reconstruction, cooperating with the adaptive sampling strategy for quick and effective densification.

\item RDG-GS attains superior results on $4$ scene-level and object-level benchmarks with higher rendering quality and real-time application speed. 
\end{itemize}

\section{Related Work}
\subsection{Novel View Synthesis.}
Synthesizing novel views~\cite{avidan1997novel} from sparse views while preserving accuracy remains a persistent challenge. 
Previous works~\cite{deng2022depthdsnerf,DBLP:journals/corr/abs-2112-03288ddpnerf,somraj2023vip,somraj2023simplenerf} has focused on the Neural Radiance Fields (NeRFs)~\cite{mildenhall2021nerf}, which learn an implicit neural representation of the scene, employing MLPs to map coordinates and using volume to
render color and density. Due to its slow training, inference speeds, and substantial computational costs, many efforts focus on enhancing efficiency~\cite{chen2022tensorf,fridovich2022plenoxels,garbin2021fastnerf,muller2022instant,SunSC22dvgo}, generation quality~\cite{barron2021mip,wang2023f2,barron2023zip,chen2022aug,guo2022nerfren,suhail2022light,wang2022clip,guedon2024sugar}, or striking a balance between the two~\cite{sun2022direct,schwarz2022voxgraf,wynn2023diffusionerf,song2023d,fridovich2023k,muller2022instant}, there still exists a significant gap between achieving real-time rendering speed and high-resolution rendering quality with photorealism.
3D Gaussian splatting (3D-GS)~\cite{kerbl20233dggs} replaces the laborious volume rendering in NeRF with efficient 3D differentiable splats, thereby rendering images with intricate shapes and appearances by representing scenes as Gaussians. 3D-GS enables real-time rendering of high-quality scenes.
While some GS-based works~\cite{shao2024splattingavatar,yang2023deformable,niedermayr2023compressed,li2024endosparserealtimesparseview,chan2024point,huang20242d} exhibit remarkable performance under dense input views, a persistent challenge persists in the form of sharp quality degradation when confronted with sparse input views.

\subsection{Sparse-view 3D Reconstruction}

Sparse-view 3D reconstruction aims to construct novel scenes with sparse input views.
Some NeRF-based endeavors~\cite{jain2021putting,cong2023enhancing,liu2020neural,yu2021pixelnerf,deng2022depthdsnerf,DBLP:journals/corr/abs-2112-03288ddpnerf,somraj2023vip,somraj2023simplenerf,wu2024reconfusion} are dedicated to pre-train models on large-scale datasets to enhance the performance. Some~\cite{yang2023freenerf,kim2022infonerf,niemeyer2022regnerf,yin2024fewviewgs} seek to constrain the specific regions by introducing regularization terms. 
CoR-GS~\cite{zhang2025cor} improves sparse-view 3DGS by leveraging point and rendering disagreements to detect and suppress reconstruction errors through co-pruning and pseudo-view regularization. However, this method cannot capture accurate geometric structures or high-frequency texture details.
Besides, numerous works~\cite{wang2023sparsenerf,uy2023scade,zhang2025cor,cheng2024sparsegnv} utilize coarse depth supervision to constrain sparse neural fields. While these works are effective, the slow training, inference speeds, and high computational costs limit their practical applications.

\begin{figure*}[htbp]
\centering
\includegraphics[width=1\linewidth]{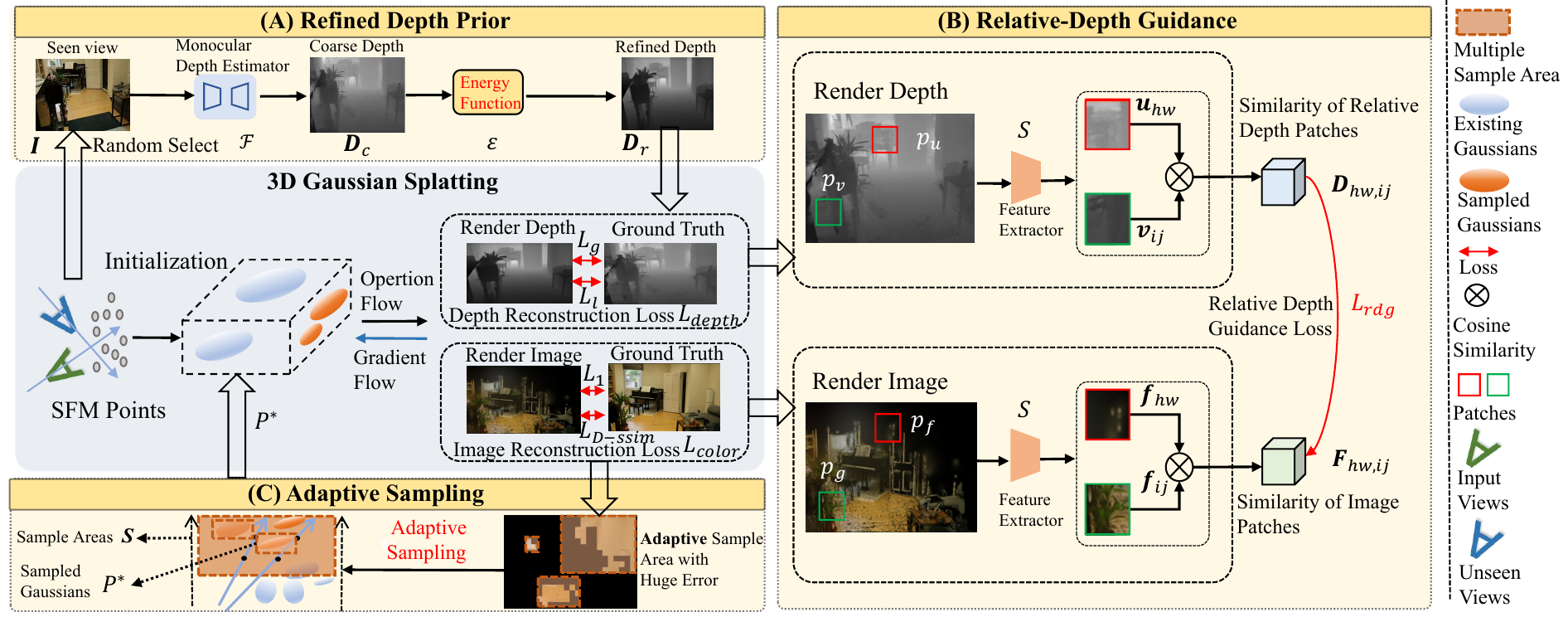}
\caption{{The network structure of RDG-GS.} (A) We obtain the refined depth by optimizing the energy module to insert global and fine-grained scene information into the optimization of Gaussian Splatting.
(B) We propose the relative depth
guidance by optimizing the similarity between spatially correlated patches of depth and images to overcome the view-inconsistent spatial information caused by the absolute depth and guide scene geometry. (C) We employ adaptive densification by sampling areas with huge training errors for more accurate and quick rendering.}
\label{main}
\end{figure*}
For recent powerful 3D GS-based works~\cite{xiong2023sparsegs,li2023spacetime,guo2024depth,zhang2025cor,huang20242d}, some utilize coarse depth from pre-trained monocular depth estimators~\cite{oquab2023dinov2,bhat2023zoedepth,birkl2023midas} for varying degrees of supervision. FSGS~\cite{zhu2023fsgs} directly optimizes the Gaussian by incorporating monocular depth priors and virtual training views. 
DNGaussian~\cite{li2024dngaussian} incorporates dual depth regularizations to refine the geometric shape of the 3D radiance field, leveraging depth priors to produce high-quality rendering results.
Notably, all the above 3D GS-based methods~\cite{zhu2023fsgs,li2024dngaussian} adopt coarse depths generated from pre-trained monocular depth estimators as ground truth for depth supervision. 
However, existing methods~\cite{zhu2023fsgs,li2024dngaussian,guo2024depth} rely solely on absolute estimated depth for geometry optimization, which frequently introduces inaccuracies that undermine the reconstruction quality of Gaussian splats. Besides, these monocular estimated depths suffer from estimation errors and introduce inconsistency problems, causing Gaussians to form blurry or incorrect shapes and thereby degrading rendering quality. 
For ours, we employ relative depth guidance by refining the spatial depth–image similarity, ensuring consistent geometric reconstruction and fine-grained refinement. Additionally, we incorporate both global and local scene information into Gaussians through refined depth priors to achieve accurate geometry and high-fidelity reconstruction.

\section{Method}
\label{method}

As illustrated in Fig.~\ref{main}, we propose RDG-GS, a novel sparse-view 3D reconstruction model with (A) refined depth priors with  consistent geometry and high-frequency details for regularization, unlike most methods~\cite{zhu2023fsgs,li2024dngaussian,guo2024depth} using coarse estimated depth supervision for image rendering, (B) relative depth guidance based on 3D spatial similarities to capture view-consistent spatial geometry information for accurate geometry rendering, and (C) adaptive sampling for densify initial Gaussians in error-prone regions to enhance rendering quality.

In this section, we overview the preliminary of 3D Gaussian Splatting~\cite{kerbl20233dggs} and the introduction of rendered image and depth in Sub-section~\ref{define}.
We provide the generation method of refined depth priors in Sub-section~\ref{3d depth} and the relative depth guidance stage in Sub-section~\ref{Global-depth Guiding}. We also describe the detail strategy of the adaptive sampling for density and training objective in Sub-section~\ref{Semantic Adaptive Sampling}.



\subsection{Preliminary and Problem Definition}
\label{define}
 Representing 3D Gaussians~\cite{kerbl20233dggs} as point clouds provide a clear depiction of 3D scenes, where each Gaussian is characterized by a covariance matrix $\Sigma$ and a centroid $\chi$, representing its mean value:
   $ G(X) = {e^{ - \frac{1}{2}{\chi ^T}{\Sigma ^{ - 1}}\chi }}$.
To facilitate differentiable optimization, the covariance matrix $\Sigma$ consists of  a scaling matrix $M$ and a rotation matrix $R$:
   $ \Sigma  = RM{M^T}{R^T}$.
3D Gaussians utilize differential splitting in camera planes to render novel views. The covariance matrix ${\Sigma} ^{'} = {J}{W}\Sigma {W^T}{J^T}$ in camera coordinates is computed using the transform matrix $W$ and the Jacobian matrix $J$ from the affine approximation of the projective transformation.

\noindent \textbf{Image Rendering.} Each 3D Gaussian is defined by several attributes: color from spherical harmonic (SH), coefficients ${\rm C}$, opacity $\alpha$, rotation $r$, scaling $s$, and position $\chi$.
To render the 2D images $\boldsymbol{I}_o$, the 3D GS arranges all the $N$ Gaussian points contributing to a pixel and combines the arranged Gaussians that overlap the pixels: 
\begin{equation}
\label{color}
    \boldsymbol{I}_o = \sum\limits_{i \in N} {{c_i}} {\alpha _i}\prod\limits_{j = 1}^{i - 1} {(1 - } {\alpha _i})
\end{equation}
where ${c_i}$ is the color computed from the spherical harmonic (SH) coefficients ${\rm C}$, and ${\alpha_i}$ is the density and then multiplied by adjustable per-point opacity and spherical harmonic color coefficients.

\noindent \textbf{Depth Rendering.} In order to realize depth regularization for geometry optimization, we enable depth back-propagation and implement the differentiable depth rasterizer by following FSGS~\cite{zhu2023fsgs} pipeline. Specifically, we make use of alpha blend rendering in 3D-GS for depth rasterization, where z-buffers of ordered gauss that contribute to pixels are accumulated to generate depth values. The rendered depth $\boldsymbol{D}_o$ can be defined as:
\begin{equation}
     \boldsymbol{D}_o = \sum\limits_{i \in N} {{d_i}} {\alpha _i}\prod\limits_{j = 1}^{i - 1} {(1 - } {\alpha _i})
\end{equation}
where $d_i$ represents the $z$-buffer of the i-th Gaussian, ${\alpha_i}$ is the density which is computed same as in Eq.~\ref{color}.

\noindent \textbf{Initialization.} Notably, 3D Gaussian Splatting~\cite{kerbl20233dggs} employs heuristic Gaussian densification based on the average gradient magnitude in view space positions exceeding a threshold. Although effective with sufficient SfM points, this approach struggles with extremely sparse point clouds from sparse views, leading to overfitting training views and poor generalization to new viewpoints.

\subsection{Refined Depth Prior}
\label{3d depth}
It's essential to supplement the geometry of the local Gaussian splats for rendering a reasonable geometric shape. We first obtain the coarse estimated depth $\boldsymbol{D}_c$ from the monocular depth estimator $\mathcal{F}$~\cite{oquab2023dinov2}. Building on this, we aim to generate the refined depth that provide correct geometry and fine-grained details.

\subsubsection{Refined Depth}
To provide accurate geometric and high-frequency texture information for 3D-GS~\cite{kerbl20233dggs}, we refined the coarse depth $\boldsymbol{D}_c$ with the guidance of high-quality images $\boldsymbol{I}$.
Inspired by depth recovery~\cite{rgb-guided}, we adopt the energy function $\mathcal{E}$ to integrate correct geometry into depth, capturing global consistency and local similarity in geometric structure, while suppressing redundant textures. 

\noindent \textbf{Energy function.} In our work, we adopt the energy function $\mathcal{E}$ to infer the pixels of the refined depth random field $\boldsymbol{D}_r$ given the coarse depth field $\boldsymbol{D}_c$ and image field $\boldsymbol{I}$. 
The target refined depth ${\boldsymbol{D}_r}$ composing $n$ pixels, is then inferred by minimizing the energy function $\mathcal{E}$, which we can denote as:
\begin{equation}
\label{energy}
    {\boldsymbol{D}_r} = \arg {\max _{{\boldsymbol{D}_r}}} \mathcal{E}({\boldsymbol{D}_r} \mid \boldsymbol{I},\boldsymbol{D}_c)
\end{equation}
The energy function $\mathcal{E}$ comprises three modules: the global structural consistency module $\psi_u$, the local similarity module $\psi_p$, and the texture detail constraint module $\psi_h$, which can be defined:
\begin{equation}
\begin{aligned} 
   \mathcal{E}(\boldsymbol{D}_r) = &\sum\nolimits_{i} \Bigl[w_u \,\psi_u(i)\,g_u^i \;+\; w_h \,\psi_h(i)\Bigr]\\
& \quad + \sum\nolimits_{i<j} w_p \,\psi_p(i,\,j)\,g_p^{(i,j)}
\end{aligned}
\end{equation}
where $w_u$, $w_p$, $w_h$ are the weights, and the $i$, $j$ are the pixels, and ${g}_u^i$, ${g}_p^{(i,j)}$ are the high-frequency weight, as detailed below.
The modules $\psi_u(i)$, $\psi_p(i, j)$, and $\psi_h(i)$, which can be defined as:
\begin{equation}
\begin{aligned}
\psi_{u}(i) & =-\log\left(\mathrm{SSIM}\left(\mathcal{S}(D_{r}(i)),\mathcal{S}(I(i))\right)\right), \\
\psi_{p}(i,j)  &= \resizebox{0.3\textwidth}{!}{$\left(1-\exp\left(-\frac{ \mid D_r(i)-D_r(j)\mid^2}{2\theta_\mu^2}\right)\right)$} \\
 & \resizebox{0.3\textwidth}{!}{$ \cdot\exp\left(-\frac{\left\|i-j\right\|^2}{2\theta_\alpha^2}-\frac{\left\|\boldsymbol{I}(i)-\boldsymbol{I}(j)\right\|^2}{2\theta_\beta^2}\right)$}, \\
\psi_{h}(i) & =\left\|\nabla\boldsymbol{I}(i)-\nabla D_{r}(i)\right\|^{2}
\end{aligned}
\end{equation}
where $\theta _\alpha$ is the standard deviation of the global Gaussian kernel, $\theta _\mu$ and $\theta _\beta$ are for the local Gaussian kernel. $\nabla$ denotes gradients, $\mathcal{S}$ is the feature extractor~\cite{oquab2023dinov2}.
Building on the framework~\cite{rgb-guided}, the $\psi_u(i)$ captures localized geometric cues by measuring the patch-based similarity around pixel $i$ between the depth map and RGB image, while $\psi_p(i,j)$ exploits both structural and pixel-wise similarities from $\boldsymbol{I}$ to preserve global structural consistency. More details are provided in the supplementary materials. Meanwhile, $\psi_h(i)$ imposes direct constraints on the high-frequency features of $\boldsymbol{D}_r$ and $\boldsymbol{I}$, ensuring these high-frequency cues align with critical geometric edges while minimizing interference from texture noise.


\noindent \textbf{High-frequency weight.}  
We propose the high-frequency weight $g$ that quantifies each pixel’s geometric edge significance, thereby facilitating more consistent capture of depth-relevant edges while suppressing extraneous textures. Concretely, we define the univariate high-frequency weight as
$g_u^i 
    = \exp\Bigl( -\frac{\|\nabla \boldsymbol{I}(i) - \nabla \boldsymbol{D}_{r}(i)\|^2}{2\tau^2} \Bigr)$
, where $\nabla$ denotes gradient-based extraction of high-frequency features, and $\tau$ controls the sensitivity to these features. We also introduce a pairwise high-frequency similarity weight $g_p^{(i,j)} 
    = \exp\Bigl( -\frac{\|\nabla \boldsymbol{I}(i) - \nabla \boldsymbol{I}(j)\|^2}{2\gamma^2} \Bigr)$ to integrate high-frequency cues while mitigating texture-induced noise, where $\gamma$ modulates the sensitivity to high-frequency gradient similarities.

\subsection{Relative Depth Guidance}
\label{Global-depth Guiding}
To incorporate the view-consistent spatial geometry information into training, we propose relative depth guidance by optimization the similarity between spatially correlated patches of depth and images to overcome the inconsistent spatial information caused by the absolute monocular estimated depth. 

\subsubsection{Relative Depth Spatial Guidance} 
Inspired by contrastive learning~\cite{khosla2020supervised}, the model tightens the mapping of latent representations for similar instances by minimizing their distances in the feature space, while distinct instances are pushed further apart. We extend this principle to our approach: for example, spatial distances between points within the same depth map are small, whereas distances between foreground and background points are larger. By leveraging the correct spatial representation of relative depth, we can optimize the Gaussian field with view-consistent spatial geometry. However, since the evaluation of variances between these two spaces depends on the distance between points, we propose a relative depth guidance to align them effectively.

Specially, given image feature $\boldsymbol{I}_o$ and corresponding depth $\boldsymbol{D}_o$ comprising $P\times P$ patches, for distant patches in the 3D scene, we employ relative depth similarity vectors $\boldsymbol{D}$ to guide image feature similarity vectors $\boldsymbol{F}$ that are further apart, and vice versa. This incorporates view-consistent spatial geometry into scene training.

We first calculate the image similarity ${\boldsymbol{F}_{hw,ij}}$, which represents the cosine similarity between the patch $\boldsymbol{f}_{hw}$ at spatial position $(h,w)$ and  patch $\boldsymbol{f}_{ij}$ at the position $(i, j)$. We also compute the relative depth space tensor ${\boldsymbol{D}_{hw,ij}}$, same as the feature tensor ${\boldsymbol{F}_{hw,ij}}$:
\begin{equation}
\begin{aligned}
    {\boldsymbol{F}_{hw,ij}} &= \frac{{\boldsymbol{f}_{hw}} \cdot {\boldsymbol{f}_{ij}}}{\|{\boldsymbol{f}_{hw}}\|\,\|{\boldsymbol{f}_{ij}}\|}, \\
    {\boldsymbol{D}_{hw,ij}} &= \frac{{\boldsymbol{u}_{hw}} \cdot {\boldsymbol{u}_{ij}}}{\|{\boldsymbol{u}_{hw}}\|\,\|{\boldsymbol{u}_{ij}}\|}.
\end{aligned}
\end{equation}
where $\boldsymbol{f}, \boldsymbol{u}\in {\mathbb R}{^{C \times H \times W}}$ are the patches generated by feature extractor~\cite{oquab2023dinov2}
$\mathcal{S}$, $(h,w)$ and $(i,j)$ are positions of pixels in the two corresponding depth patch $\boldsymbol{u}$.

\subsubsection{Relative Depth Guidance Loss} 
For two patches in relative depth representing spatial geometry at different image locations, we first compute the similarity between depth patches to capture their spatial relationships in 3D space. This similarity quantifies how close the patches are in spatial geometry. Next, we calculate the similarity between the corresponding image patches in feature space, which reflects the visual similarity in terms of texture, color, or shape.
Thus we employ the relative depth guidance loss, which minimizes the discrepancy between image feature similarity and depth values, ensuring view-consistent spatial alignment between the depth map and image features.

Specifically, we minimize the  $L_{rdg}$ to encourage $\boldsymbol{F}_{hw,ij}$ to increase when $\boldsymbol{D}_{hw,ij}-b$ are positive and decrease when $\boldsymbol{D}_{hw,ij}-b$ are negative. Thus, the feature vector $\boldsymbol{F}_{hw,ij}$ is encouraged to match the depth vector $\boldsymbol{D}_{hw,ij}$ and obtain the correct relative geometric information.
$L_{rdg}$ is as follows:
\begin{equation}
\label{dfg_loss}
   \resizebox{0.42\textwidth}{!}{$ {L_{{{rdg}}}} = \sum\limits_{hw,ij} {\log (1 + {\text{exp}^{ - ({\boldsymbol{D}_{hw,ij}} - b)\max ({\boldsymbol{F}_{hw,ij}},0)}})} $ }
\end{equation}
where $(h,w)$ and $(i,j)$ are positions of pixels, $b$ serves as a bias for preventing collapsing, which is adaptive $b(t) = b {(t - 1)^{\mid\frac{t}{m}\mid}}$ with the $m$ training steps to capture the most significant global structure information. We also adopt zero-clamping to delete the weakly-guidance features and improve stability.

\subsection{Adaptive Sampling }
\label{Semantic Adaptive Sampling}
\subsubsection{Adaptive Sampling Strategy}
To tackle the issue of insufficient Gaussians in the initial camera distribution, which hampers convergence to high rendering quality, we introduce an adaptive sampling strategy to re-optimize point cloud initialization for Gaussian splitting. Different from the simple error threshold used in SpaceGaussian~\cite{li2023spacetime}, we first identify areas $S={s_1,...,s_M}$ with huge 3D depth training errors during the training, 
Specifically, if a patch's 
depth regularization loss $I_{{s_i}}^e$ exceeds the threshold $I_{{s_i}}^{\text{threshold}}$, then that patch is seems as the region which has a huge training error. 
The threshold is defined as the mean loss over all patches.
Thus, after the training loss stabilizes, we sample new Gaussians patch-wise along the pixel rays within these $M$ areas to prioritize areas with substantial errors rather than outlier pixels. Then, we sample rays from the center pixels of each selected patch with significant errors. Next, we uniformly sample new Gaussians within the depth range along the rays, and the resampled Gaussians $P^*$ are reintroduced into the initialization for point cloud densification. 
\begin{equation}
   P^* = P \cup \bigcup_{{s_i} \in S} \mathcal{F}_s\left(\{S_i \mid I_{{s_i}}^e > I_{{s_i}}^{\text{threshold}}\}\right)
\end{equation}
where $P$ represents the set of initial Gaussians sampled within a predefined depth range along the rays, $\mathcal{F}_s$ is the adaptive sampling strategy along the rays, and  $\cup$ represents the union operation. 

We dynamically adjust the sampling of different areas $S$ in each training iteration by resampling patches where the training loss of error area $I^e$ exceeds the average loss $I^{\text{threshold}}$ across all patches. As the loss decreases during training, the erroneous patches diminish, ultimately resulting in high-quality scenes with minimal error areas. 
It is noteworthy that adaptive sampling encourages Gaussian splats to render boundary error regions more rapidly, thereby enhancing the overall reconstruction speed. 

\subsubsection{Training objective}
Overall, the total training objective $L_{total}$ consists of three parts: the color reconstruction loss $L_{color}$, the refined depth regularization loss $L_{depth}$, and the relative depth guidance loss $L_{rdg}$. 

\noindent\textbf{Image Reconstruction Loss.} Following the 3D Gaussian Splatting~\cite{kerbl20233dggs}, the color reconstruction loss $L_{color}$ comprises a combination of the $L_1$ reconstruction loss and  D-SSIM term between the rendered image $\boldsymbol{I}_o$ and the ground truth $\boldsymbol{I}_g$:
\begin{equation}
    L_{color}={{L_1}({\boldsymbol{I}_o},{\boldsymbol{I}_g}) + \beta {L_{D - SSIM}}({\boldsymbol{I}_o},{\boldsymbol{I}_g})}
\end{equation}
where $\beta$ is a hyperparameter for balancing.

\noindent\textbf{Refined Depth Loss.} 
To obtain the refined depth $\boldsymbol{D}_r$ with accurate geometric structure and texture details, we design a refined depth loss $L_{depth}$, which comprises a global geometric consistency loss $L_g$ and a local fine-texture alignment loss $L_l$. This loss regularizes Gaussian primitives to conform to the correct geometric structure while capturing local details.

For the loss $L_g$, motivated by the FSGS~\cite{zhu2023fsgs}, we optimize the Pearson correlation between the refined depth $\boldsymbol{D}_r$ and the depth map $\boldsymbol{D}_o$ rendered by the Gaussian model to ensure consistency in the global structural distribution, mitigating the scale ambiguity issue. The loss $L_g$ is computed as:
\begin{equation}
   {L_g(\boldsymbol{D}_r,\boldsymbol{D}_o)=\left\| {\frac{{{\mathop{\rm Cov}\nolimits} \left( {{\boldsymbol{D}_{{\rm{r}}}},{\boldsymbol{D}_{{\rm{o}}}}} \right)}}{{\sigma \left( {{\boldsymbol{D}_{{\rm{r}}}}} \right) \cdot \sigma \left( {{\boldsymbol{D}_{{\rm{o}}}}} \right)}}} \right\|_1}
\end{equation}
where Cov represents covariance, $\sigma$ represents standard deviation, and $\left\| \cdot \right\|_1$ is the L1 norm.

For sparse-view inputs, global regularization can capture the overall geometric structure, but it overlooks fine local details. This leads to fluctuating noise in the Gaussian radiance field and results in poor reconstructions. Therefore, we design a loss $L_l$ specifically optimized for local fine structure. Specifically, we divide the refined depth $\boldsymbol{D}_r$ and the rendered depth $\boldsymbol{D}_o$ into patches. For each pixel $\mathbf{x}$ in the depth map, we subtract the mean of all pixels in the patch $ \boldsymbol{p}$ and divide by the standard deviation. The normalized depth can be expressed as: $\mathcal{\boldsymbol{D}}^{N}(\mathbf{x})=\frac{\mathcal{\boldsymbol{D}}(\mathbf{x})-\operatorname{mean}(\mathcal{\boldsymbol{D}}(\boldsymbol{p}))}{\operatorname{std}(\mathcal{\boldsymbol{D}}(\boldsymbol{p}))+\epsilon}$
where $\epsilon$ is a numerical stability value. From this, we can calculate the optimization loss for local details through $L_2$ normalization, which can be represented as:
\begin{equation}
    L_l(\boldsymbol{D}_r,\boldsymbol{D}_o)=L_2(\boldsymbol{D}_r
^N,\boldsymbol{D}_o^N)
\end{equation}

The final loss of the refined depth can be formulated by:
\begin{equation}
    L_{depth}=L_g(\boldsymbol{D}_r,\boldsymbol{D}_o)+\lambda L_l(\boldsymbol{D}_r,\boldsymbol{D}_o)
\end{equation}
where $\lambda$ is the hyperparameter for balancing.

\noindent\textbf{Total Loss.} The total loss can be formulated  as follows:
\begin{equation}
       {L_{total}} = L_{color} +  {L_{depth}}+ \omega {L_{rdg}}
\end{equation}
where $\beta$, and $\omega$ are the loss parameters, $\omega$ is also adaptive  $\omega (t) = \omega {(t - 1)^{\mid\frac{t}{m}\mid}}$ with every $m$ training steps.

\section{Experiment}
\subsection{Datasets and Implementation Details}

\subsubsection{Datasets}
We evaluate our model on $4$ scene-level and object-level datasets.

\noindent \textbf{Mip-NeRF360}~\cite{barron2022mip260}  comprise $9$ unbounded indoor and outdoor scenes. Following the official setting, we utilize 24 viewpoints from the $7$ scenes for comparison and training, with images downsampled to 1/2, 1/4, and 1/8 resolutions. The selection of test images follows the same protocol as that of the LLFF~\cite{mildenhall2019localllff}. To the best of our knowledge, we are pioneering the exploration of novel view synthesis within unbounded scenes in Mip-NeRF360~\cite{barron2022mip260} with sparse-view inputs. 
\textbf{NeRF-LLFF~\cite{mildenhall2019localllff}} consists of 8 complex scenes captured with a frontal-facing camera. We adhere to the official training/testing setup. 
We randomly choose 10 seed particles and average the results over 10 experiments. Following FSGS~\cite{zhu2023fsgs}, 
we train using $3$ views and evaluate $8$ views under resolutions of $1008\times 756$ and $504 \times 378$.

\noindent\textbf{DTU~\cite{6909453DTU}} comprises $124$ object-centric scenes captured by a set of fixed cameras. Following SparseNeRF~\cite{wang2023sparsenerf} and RegNeRF~\cite{niemeyer2022regnerf}, we adopt $15$ sample scenes, each containing $3$ training views and $15$ test views, all of which undergo a 4× downsampling. 

\noindent\textbf{Blender~\cite{mildenhall2021nerf}} comprises 8 photorealistic synthetic object images synthesized using Blender. Following DietNeRF~\cite{jain2021puttingdietnerf}, we train with 8 views and test with 25 views. Throughout the experimentation, all images are downsampled by a factor of 2 to dimensions of 400 × 400.

\subsubsection{Implementation Details}
Following 3D-GS~\cite{kerbl20233dggs}, we train our model on the single NVIDIA 3090 GPU with the Pytorch and obtain the camera parameters and sparse depth from the COLMAP SfM~\cite{schonberger2016structureSFM}. The coarse depth is generated by monocular depth estimator DPT~\cite{ranftl2021visiondpt}. Following DNGaussian~\cite{li2024dngaussian}, we incorporated depth into the CUDA kernel for rasterization and re-registered it. 
We set the total iterations $m$ to $6000$ and we apply the depth regularization after $1000$ interactions, and the densification interval is set to $100$. 

\begin{table*}
	\centering
  \small
 \caption{Comparisons between RDG-GS and SOTA methods on Mip-NeRF360~\cite{barron2022mip260} with $24$ training views. All the works are optimized per scene. We color the top-3 results with different colors, which are the \textbf{\colorbox{top1}{best}}, \textbf{\colorbox{top2}{second best}}, and \textbf{\colorbox{top3}{third best}}. },
	\scalebox{0.85}{
	\setlength{\tabcolsep}{0.1pt}
\begin{tabular}{lccccccccccccc}
\hline
\multicolumn{1}{c}{\multirow{2}{*}{Methods}} & \multirow{2}{*}{Type}                                                                           & \multicolumn{4}{c|}{1/2 Resolution}                                                & \multicolumn{4}{c|}{1/4 Resolution}                                                & \multicolumn{4}{c}{1/8 Resolution}                               \\  
\multicolumn{1}{c}{}                         &                                                                                                 & PSNR↑          & SSIM↑          & LPIPS↓           & \multicolumn{1}{c|}{RMSE↓ }         & PSNR↑          & SSIM↑          & LPIPS↓                & \multicolumn{1}{l|}{RMSE↓}          & PSNR↑          & SSIM↑         & LPIPS↓         & RMSE↓          \\ \hline
Mip-NeRF360~\cite{barron2022mip260}   & \multirow{3}{*}{\begin{tabular}[c]{@{}c@{}}SOTA \\ NeRF-based\end{tabular}}                                & 17.83          & 0.451          & 0.557               & \multicolumn{1}{l|}{2.386}          & 19.78          & 0.530           & 0.431            & \multicolumn{1}{l|}{1.983}          & 21.23          & 0.613          & 0.351             & 1.578          \\
DietNeRF~\cite{jain2021puttingdietnerf}          &                           & 16.56          & 0.381          & 0.543               & \multicolumn{1}{l|}{2.281}          & 19.11          & 0.482          & 0.452               &\multicolumn{1}{l|}{1.821}          & 20.21          & 0.557          & 0.387              & 1.524          \\
RegNeRF~\cite{niemeyer2022regnerf}             &                         & 18.14          & 0.458          & 0.502             &\multicolumn{1}{l|}{ 2.136 }         & {20.55}          & 0.546          & 0.398               & \multicolumn{1}{l|}{1.774}          & 22.19          & 0.643          & 0.335             & 1.519          \\ \hline
FreeNeRF~\cite{yang2023freenerf}     & \multirow{2}{*}{\begin{tabular}[c]{@{}c@{}}SOTA NeRF-based \\ for sparse-view\end{tabular}}                                & 18.35          & 0.471          & 0.481                 & \multicolumn{1}{l|}{2.081}         & {21.39}          & 0.587          & \cellcolor{top3}0.377             & \multicolumn{1}{l|}{1.692}          & 22.78          & 0.689          & 0.323            & 1.487          \\
SparseNeRF~\cite{wang2023sparsenerf}    &                               & \cellcolor{top3}19.02          & \cellcolor{top3}0.497          & \cellcolor{top3}0.476                & \multicolumn{1}{l|}{\cellcolor{top3}2.013}          & {\cellcolor{top3}21.43}          & \cellcolor{top3}0.604          & 0.389              & \multicolumn{1}{l|}{\cellcolor{top3}1.631}          & \cellcolor{top3}22.85          & \cellcolor{top3}0.693          & \cellcolor{top3}0.315             & 1.469          \\ \hline
3D-GS~\cite{kerbl20233dggs}                 & \multirow{3}{*}{\begin{tabular}[c]{@{}c@{}}SOTA 3D GS-based\\ for sparse-view\end{tabular}}                                  & 17.12          & 0.476          & 0.514                  & \multicolumn{1}{l|}{2.124}          &{19.93}          & 0.588          & 0.401          &\multicolumn{1}{l|}{ 1.682 }         & 20.89          & 0.633          & 0.317                & \cellcolor{top3}1.422          \\
FSGS~\cite{zhu2023fsgs}                      &        & \cellcolor{top2}20.11          & \cellcolor{top2}0.511          & \cellcolor{top2}0.414                & \multicolumn{1}{l|}{\cellcolor{top2}1.982 }         &{\cellcolor{top2}22.52}          & \cellcolor{top2}0.673          & \cellcolor{top2}0.313                  & \multicolumn{1}{l|}{\cellcolor{top2}1.523}          & \cellcolor{top2}23.70          & \cellcolor{top2}0.745          & \cellcolor{top2}0.230             & \cellcolor{top2}1.388          \\
CoR-GS~\cite{zhang2025cor}              &              &-       &-        &-                & \multicolumn{1}{c|}{- }         &{-}          & -    & -              & \multicolumn{1}{c|}{-}          & 23.39          & 0.727          & 0.271                &-         \\ \hline
\textbf{Ours}                      & \multicolumn{1}{l}{{GS-based  for sparse-view}}           & \textbf{\cellcolor{top1}22.67} & \textbf{\cellcolor{top1}0.548} & \textbf{\cellcolor{top1}0.354}  & \multicolumn{1}{l|}{\textbf{\cellcolor{top1}1.731}} &{\textbf{\cellcolor{top1}25.01}} & \textbf{\cellcolor{top1}0.738} & \textbf{\cellcolor{top1}0.245}& \multicolumn{1}{l|}{\textbf{\cellcolor{top1}1.342} }& \textbf{\cellcolor{top1}26.03} & \textbf{\cellcolor{top1}0.794} & \textbf{\cellcolor{top1}0.219} & \textbf{\cellcolor{top1}1.301} \\ \hline
\end{tabular}}
	\label{Mip-NERF}
\end{table*}

\begin{table*}[!htp]
	\centering
 \caption{The evaluation results of our method, compared with other advanced approaches~\cite{kerbl20233dggs,zhu2023fsgs,zhang2025cor} on the Mip-NeRF360 dataset~\cite{somraj2023simplenerf}, using 12 and 24 training views.} 
      \setlength{\tabcolsep}{16pt}
	\scalebox{0.92}{
\begin{tabular}{lcccccc}
\hline
\multicolumn{1}{c}{\multirow{2}{*}{Method}} & \multicolumn{3}{c|}{{12-view}} & \multicolumn{3}{c}{24-view} \\
\multicolumn{1}{c}{}                        & PSNR↑   & SSIM↑   & \multicolumn{1}{c|}{LPIPS↓}  & PSNR↑   & SSIM↑   & LPIPS↓  \\ \hline
3DGS~\cite{kerbl20233dggs}                                        & 18.52   & 0.523   & \multicolumn{1}{c|}{\cellcolor{top2}0.415}   & 22.80   & 0.708   & 0.276   \\
FSGS~\cite{zhu2023fsgs}                                        & \cellcolor{top3}18.80   & \cellcolor{top3}0.531   & \multicolumn{1}{c|}{\cellcolor{top3}0.418}   & \cellcolor{top3}23.28   & \cellcolor{top3}0.715   & \cellcolor{top3}0.274   \\
CoR-GS~\cite{zhang2025cor}                                      & \cellcolor{top2}19.52   & \cellcolor{top2}0.558   & \multicolumn{1}{c|}{\cellcolor{top3}0.418}   & \cellcolor{top2}23.39   & \cellcolor{top2}0.727   & \cellcolor{top2}0.271   \\ \hline
\textbf{Ours}                               & \textbf{\cellcolor{top1}21.67}   &\textbf{\cellcolor{top1} 0.596}   & \multicolumn{1}{c|}{\textbf{\cellcolor{top1}0.396}}   & \textbf{\cellcolor{top1}26.03}  & \textbf{\cellcolor{top1}0.794}   & \textbf{\cellcolor{top1}0.219}   \\ \hline
\end{tabular}
}
	\label{training_views_mip}
\end{table*}

The loss term parameters $\beta$, $\lambda$, and original $\omega$ are set to $0.4$, $0.1$, and $0.05$, respectively. The initial value of $b$ in Eq.~\ref{dfg_loss} is set to $0.4$. The $\theta _\alpha$, $\theta _\mu$, and $\theta _\beta$ of coarse and fine-grained modules $\psi_p$ are set to $35$, $10$, $10$ and $10$, $2$, $2$, respectively. 
The parameter $\tau$ controlling the sensitivity to high-frequency features sets to 5, and $\gamma$ which controls the sensitivity to high-frequency gradient similarity sets to 10, respectively. The numerical stability value $\epsilon$ sets to $10^{-6}$.

Employing the original settings of 3D Gaussian Splatting~\cite{kerbl20233dggs}, we constructed the model from unstructured sparse-view images and employed Structure-from-Motion~\cite{schonberger2016structureSFM} (SfM) for image calibration. We conducted dense stereo matching under COLMAP using ``patch match Stereo" and utilized stereo fusion to merge the resulting 3D point clouds. Next, we initialized the positions of 3D Gaussian splats based on the fused point cloud. For the feature extraction $\mathcal{S}$, we employed the DINO model~\cite{oquab2023dinov2} with a patch size set to 8, granularity set to 1, embedding layer dimension of 512, and a random crop ratio of 0.5. For the Gaussian splats,
\begin{figure}[!htbp]
\centering
\includegraphics[width=1\linewidth]{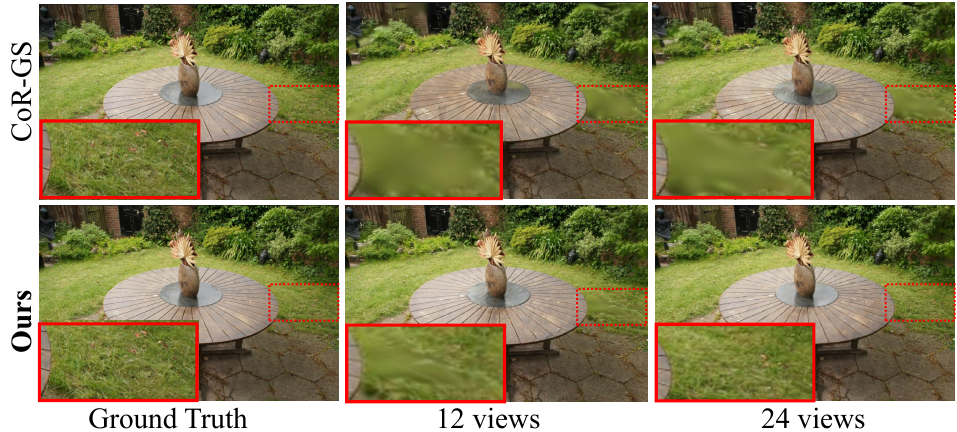}
\caption{Visual comparisons of different 12, 24 training views of RDG-GS (ours) and CoR-GS~\cite{zhang2025cor} on Mip-NeRF360~\cite{barron2022mip260}.  }
\label{views_viz_mip}
\end{figure}
we set the spherical harmonics (SH) to 2. We initialized opacity to 0.1 and adjusted the scale to match the average distance between points. Following the methodology of FSGS~\cite{zhu2023fsgs}, we set the learning rates for position, opacity, scale, and rotation to 0.0002, 0.003, 0.06, 0.005, and 0.002 respectively. At iterations of 1000 and 3000, all the opacities of Gaussian splats were reset to 0.04 to eliminate low-opacity artifacts.

\subsubsection{Metrics}
We evaluate performance through 4 metrics, including the PSNR, SSIM, LPIPS, and RMSE, with the detailed computational methods elaborated in the following sections.

\noindent \textbf{PSNR}
Peak Signal-to-Noise Ratio (PSNR) is commonly used to measure the quality of reconstructed or compressed images. It quantifies the difference between the original and the reconstructed images in terms of peak signal power and noise.

\noindent \textbf{SSIM}
The structural Similarity Index (SSIM) is a metric used to assess the similarity between two images by considering their perceived structural information. Unlike traditional metrics like Mean Squared Error (MSE), SSIM takes into account the perceived changes in structural information, luminance, and contrast that are important for human perception.
Following FreeNeRF~\cite{yang2023freenerf} and RegNeRF~\cite{niemeyer2022regnerf}, we use the ``structural similarity" API in scikit-image to calculate the SSIM score.

\begin{figure*}[!htbp]
\centering
\includegraphics[width=1\linewidth]{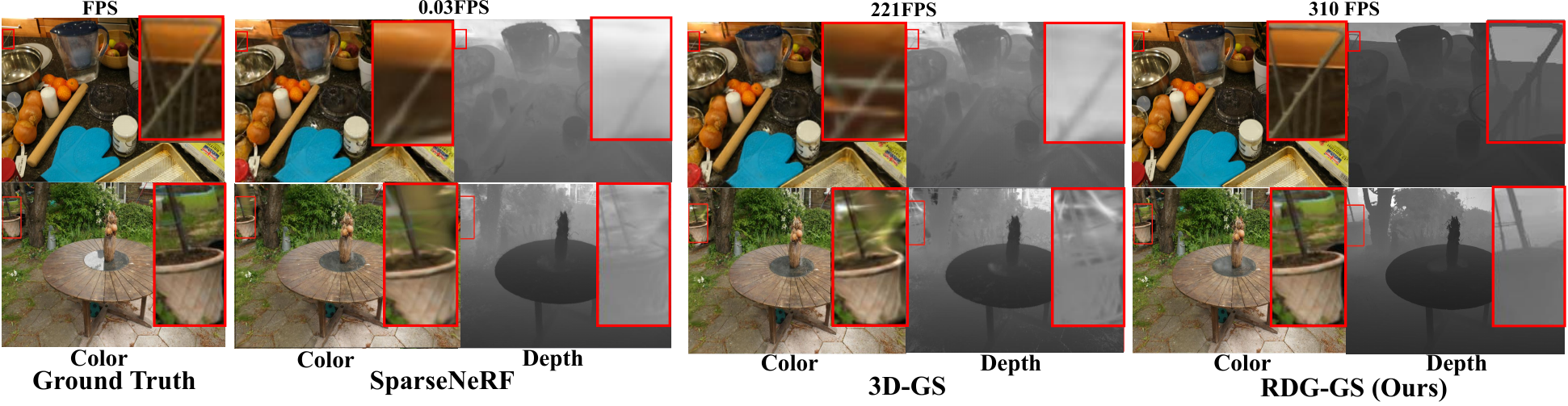}
\caption{Comparison of RDG-GS with the SOTA works SparseNeRF~\cite{wang2023sparsenerf} and 3D Gaussian Splatting~\cite{kerbl20233dggs} of sparse-view 3D reconstruction with $24$ training views. The proposed RDG-GS has super outperformance in refined depth priors with correct geometric shapes and fine-grained details, as well as the real-time 3D reconstruction of high-quality scenes.}
\label{fig:teaser}
\end{figure*}
\begin{figure*}[htbp]
\centering
\includegraphics[width=1\linewidth]{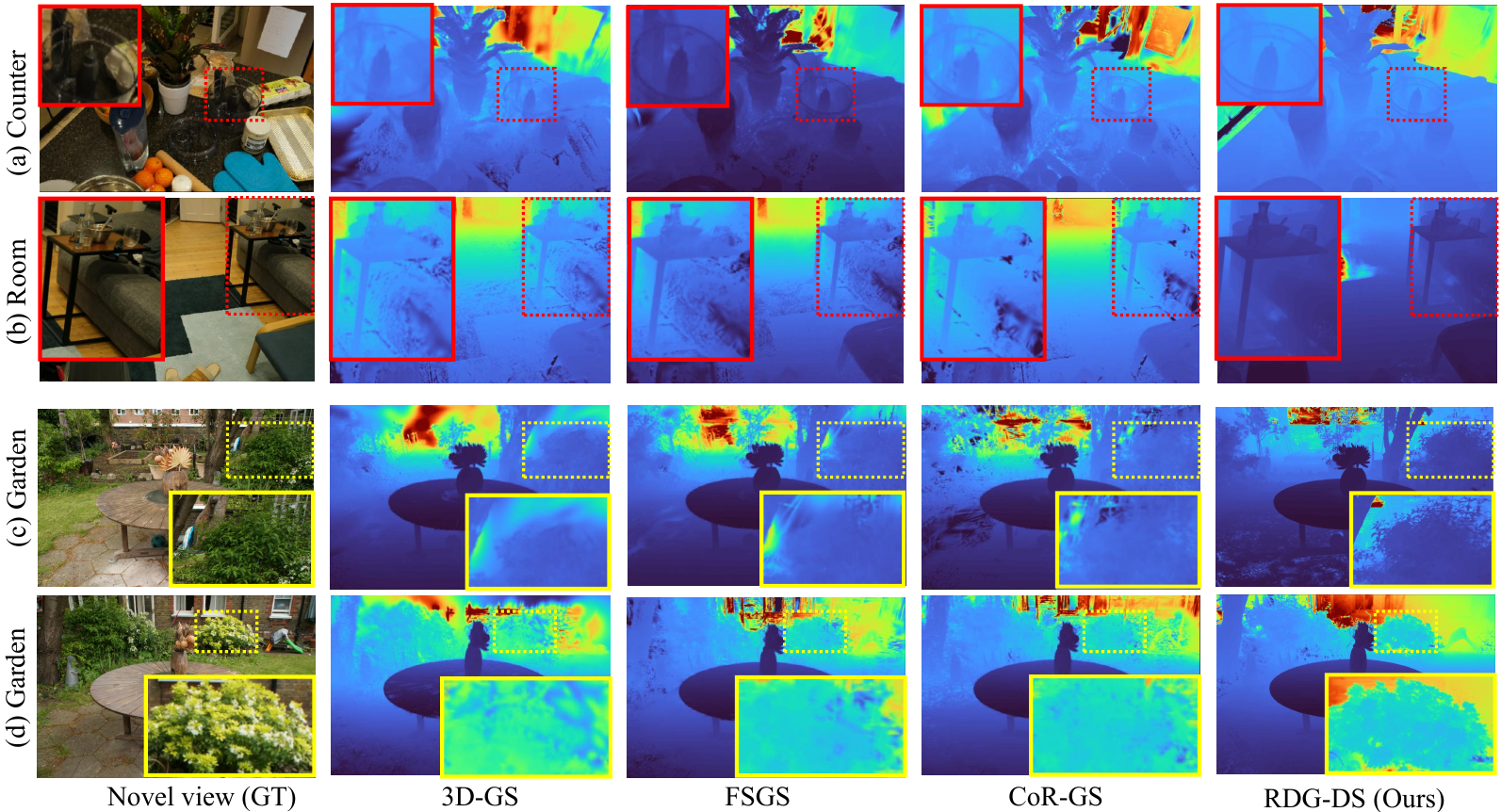}
\caption{More qualitative results of rendered depth in Mip-NeRF360 dataset~\cite{barron2022mip260} between RDG-GS, 3D-GS ~\cite{kerbl20233dggs}, CoR-GS~\cite{zhang2025cor}, and FSGS~\cite{zhu2023fsgs} in generating accurate geometric scenes and high-frequency texture details. }
\label{more_depth}
\end{figure*}

\noindent \textbf{LPIPS}
The Learned Perceptual Image Patch Similarity (LPIPS) metric serves as a perceptual similarity measure designed to assess the visual resemblance between two images in a manner that aligns with human perception. Unlike traditional evaluation metrics such as PSNR or SSIM, which rely on predefined formulas and handcrafted features, LPIPS is data-driven and leverages learned features to achieve a more perceptually relevant evaluation. In line with the approach employed by FreeNeRF~\cite{yang2023freenerf}, we utilize a pre-trained AlexNet model to calculate the LPIPS score, ensuring an effective measure of perceptual similarity. 

\noindent \textbf{RMSE}
Root Mean Square Error (RMSE) is computed through $\sqrt {\frac{1}{N}{{\sum\limits_{{\rm{i}} = 1}^N {\mid {x_i} - x_i^*\mid^2} }}} $, serves as a statistical metric that gauges how well a model's predictions align 
with actual values by quantifying deviations between predicted and observed data points. This single value succinctly encapsulates the model's overall prediction 
discrepancy, offering a straightforward yet powerful means of evaluation. 

\begin{figure*}[!htbp]
\centering
\includegraphics[width=1\linewidth]{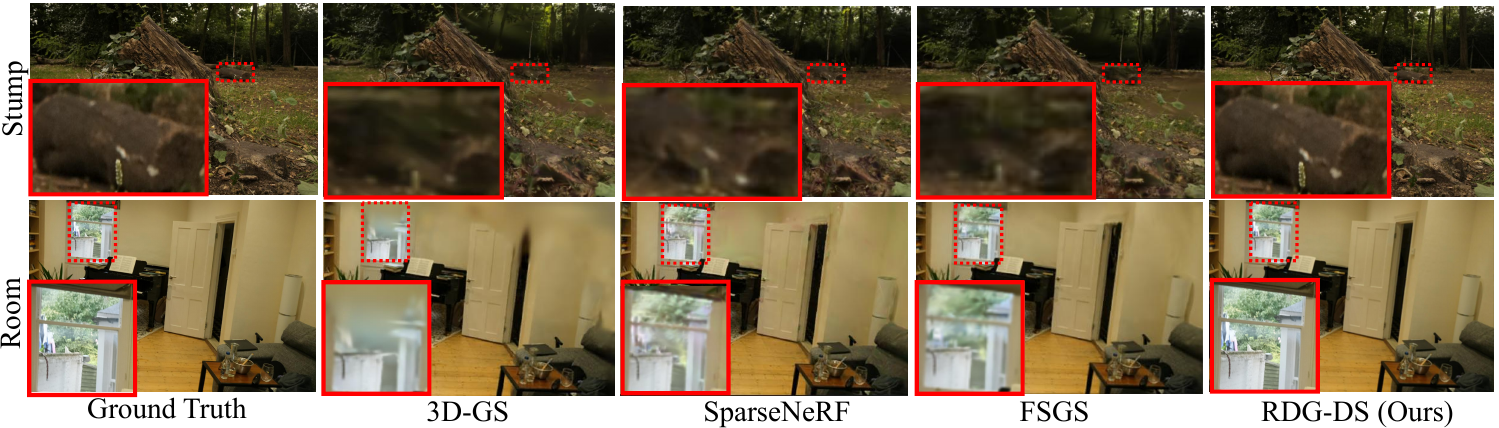}
\caption{Qualitative results in Mip-NeRF360~\cite{somraj2023simplenerf} dataset (1/4×) between Ours, 3D-GS~\cite{kerbl20233dggs}, SparseNeRF~\cite{wang2023sparsenerf}, and FSGS~\cite{zhu2023fsgs}. }
\label{viz_360}
\end{figure*}

\begin{figure*}[!htbp]
\centering
\includegraphics[width=1\linewidth]{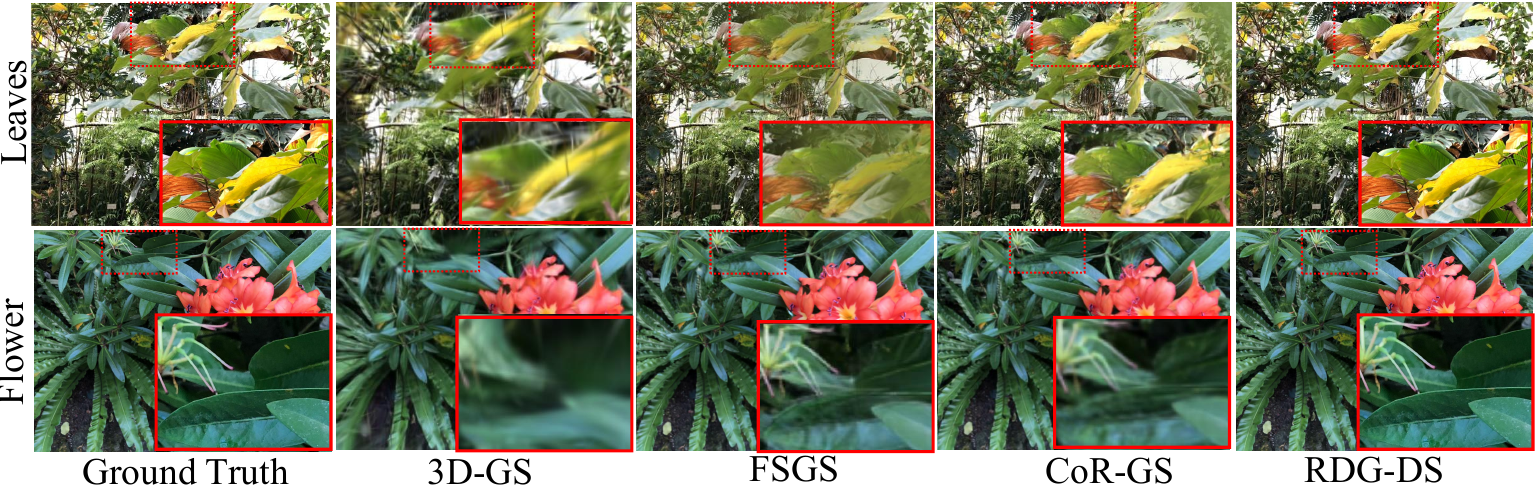}
\caption{Qualitative comparison in NeRF-LLFF~\cite{mildenhall2019localllff} dataset between our model and the 3D-GS~\cite{kerbl20233dggs}, FSGS~\cite{zhu2023fsgs}, and CoR-GS~\cite{zhang2025cor} works. }
\label{viz_llff}
\end{figure*}

\begin{figure*}[!htbp]
\centering
\includegraphics[width=1\linewidth]{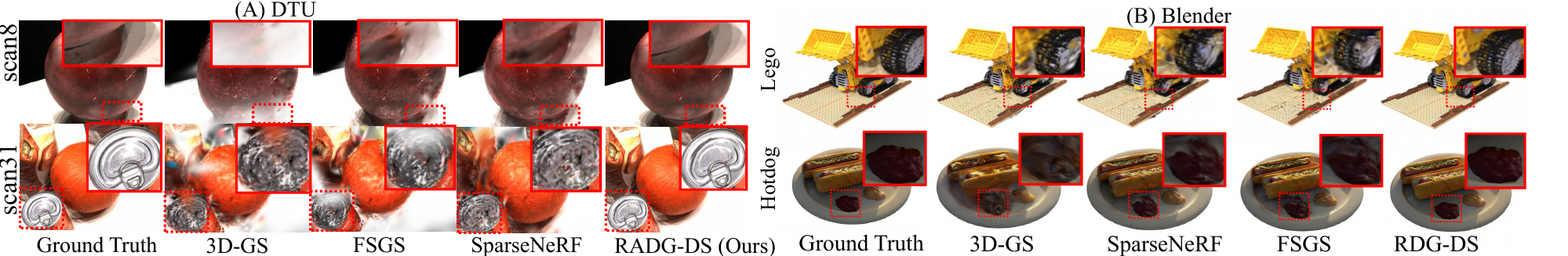}
\caption{Qualitative comparison in DTU~\cite{6909453DTU} and Blender~\cite{mildenhall2021nerf} datasets between RDG-GS and the 3D-GS~\cite{kerbl20233dggs}, SparseNeRF~\cite{wang2023sparsenerf}, and the FSGS~\cite{zhu2023fsgs}. }
\label{viz_blender}
\end{figure*}
\subsection{Experiments and Results}
\subsubsection{Comparison on Mip-NeRF360}


As demonstrated in Table~\ref{Mip-NERF}, our approach achieves state-of-the-art performance across various resolution settings, significantly outperforming both the leading NeRF-based~\cite{yang2023freenerf,wang2023sparsenerf} and 3D-GS-based methods~\cite{kerbl20233dggs,zhu2023fsgs,zhang2025cor} under sparse-view configurations. Remarkably, our model maintains competitive results, even when compared to SOTA NeRF-based methods~\cite{barron2022mip260,jain2021puttingdietnerf,niemeyer2022regnerf} under dense-view settings.
Combined with Table~\ref{work_transfer}, compared to SOTA NeRF-based models~\cite{wang2023sparsenerf,yang2023freenerf}, our model achieves over $4000\times$ faster real-time rendering while maintaining high-quality images with detailed results.

\begin{table*}[!htbp]
	\centering
 \caption{The comparisons between RDG-GS and SOTA methods on DTU~\cite{6909453DTU} with $3$ training views and Blender~\cite{mildenhall2019localllff} datasets with $8$ training views.} 
	\scalebox{0.85}{
	\setlength{\tabcolsep}{10pt}
\begin{tabular}{lcccccccccc}
\hline
\multicolumn{1}{c}{\multirow{2}{*}{Methods}} & \multicolumn{4}{c|}{DTU}                                                                              & \multicolumn{4}{c}{Blender}                                                         \\ 
\multicolumn{1}{c}{}        & PSNR↑          & SSIM↑          & LPIPS↓         & \multicolumn{1}{c|}{RMSE↓}               & PSNR↑          & SSIM↑          & LPIPS↓         & RMSE↓           \\ \hline
Mip-NeRF360~\cite{barron2022mip260}   
& 8.68 & 0.571 & 0.353 & \multicolumn{1}{c|}{1.314}  & 20.89 & 0.830  & 0.168  & \multicolumn{1}{c}{1.244}\\
DietNeRF~\cite{jain2021puttingdietnerf}  
& 11.85 & 0.633 & 0.314  & \multicolumn{1}{c|}{1.131}& 22.50  & 0.823 & {0.124}  & \multicolumn{1}{c}{1.114} \\
RegNeRF~\cite{niemeyer2022regnerf} 
& 18.89 & 0.745 & 0.190  & \multicolumn{1}{c|}{1.021} & 23.86 & 0.852 & 0.105  & \multicolumn{1}{c}{0.941}\\
MVSNeRF~\cite{chen2021mvsnerf}
& 18.54 & 0.769 & 0.197  & \multicolumn{1}{c|}{0.482} & 24.33 & 0.881 &  {0.099}  & \multicolumn{1}{c}{\cellcolor{top3}0.901}\\   \hline
FreeNeRF~\cite{yang2023freenerf}    
& \cellcolor{top3}19.92 & 0.787 & 0.182  & \multicolumn{1}{c|}{0.971}  & 24.26 & 0.883 & 0.098  & \multicolumn{1}{c}{0.912} \\
SimpleNeRF~\cite{somraj2023simplenerf}  
&16.25& 0.751 & 0.249  & \multicolumn{1}{c|}{-} &- & - & - & \multicolumn{1}{c}{-}\\ 
SparseNeRF~\cite{wang2023sparsenerf}  
& 19.55 & 0.769 & 0.201  & \multicolumn{1}{c|}{0.203} & 24.04 & 0.876 & 0.113  & \multicolumn{1}{c}{1.107}\\ 
ReconFusion~\cite{wu2024reconfusion}  
&20.74 & 0.798 & 0.124  & \multicolumn{1}{c|}{-} &- & - & - & \multicolumn{1}{c}{-}\\ \hline
3D-GS~\cite{kerbl20233dggs}
&10.99 &0.585 & 0.313 & \multicolumn{1}{c|}{\cellcolor{top2}0.118} & 21.56& 0.847 & 0.130   & \multicolumn{1}{c}{1.128} \\
DNGaussian~\cite{li2024dngaussian}
&  18.91 & 0.790 & 0.176 & \multicolumn{1}{c|}{0.179}& 24.31 & 0.886 & \cellcolor{top3}0.088   & \multicolumn{1}{c}{1.011} \\
FSGS~\cite{zhu2023fsgs}        
&  \cellcolor{top2}21.21 & \cellcolor{top3}0.782 & \cellcolor{top3}0.172  & \multicolumn{1}{c|}{\cellcolor{top3}0.161} & \cellcolor{top2}24.64 & \cellcolor{top3}0.895 & 0.095  & \multicolumn{1}{c}{\cellcolor{top2}0.883}   \\ 
CoR-GS~\cite{zhang2025cor}        
& 19.21 & \cellcolor{top2}0.853 & \cellcolor{top2}0.119  & \multicolumn{1}{c|}{-} &  \cellcolor{top3}24.43 & \cellcolor{top2}0.896 & \cellcolor{top2}0.084  & \multicolumn{1}{c}{-}   \\ \hline
\textbf{Ours}                                & \textbf{\cellcolor{top1}23.51} & \textbf{\cellcolor{top1}0.881} & \textbf{\cellcolor{top1}0.113}   & \multicolumn{1}{c|}{\textbf{\cellcolor{top1}0.092}}  & \multicolumn{1}{c}{\textbf{\cellcolor{top1}26.21}}& \textbf{\cellcolor{top1}0.907} & \textbf{\cellcolor{top1}0.074}  & \multicolumn{1}{c}{\textbf{\cellcolor{top1}0.769}}  \\ \hline
\end{tabular}}
\label{scannet and bender}
\end{table*}

\begin{table*}[!htbp]
	\centering
 \caption{The comparisons between RDG-GS and SOTA methods on NeRF-LLFF dataset~\cite{mildenhall2019localllff} with $3$ training views.} 
	\scalebox{0.85}{
	\setlength{\tabcolsep}{10pt}
\begin{tabular}{lcccccccccc}
\hline
\multicolumn{1}{c}{\multirow{2}{*}{Methods}}  & \multicolumn{4}{c|}{503 × 381 Resolution}                                                                              & \multicolumn{4}{c}{1006 × 762 Resolution}                                                         \\ 
\multicolumn{1}{c}{}        & PSNR↑          & SSIM↑          & LPIPS↓         & \multicolumn{1}{c|}{RMSE↓}                 & PSNR↑          & SSIM↑          & LPIPS↓         & RMSE↓           \\ \hline
Mip-NeRF360\cite{barron2022mip260}
& 16.11          & 0.401          & 0.460           & \multicolumn{1}{c|}{1.845}            & 15.22          & 0.351          & 0.540                   & \multicolumn{1}{l}{2.121}     \\
DietNeRF~\cite{jain2021puttingdietnerf}  
& 14.94          & 0.370           & 0.496          &  \multicolumn{1}{c|}{1.735}          & 13.86          & 0.305          & 0.578               &\multicolumn{1}{l}{ 2.073}            \\
RegNeRF~\cite{niemeyer2022regnerf} 
& 19.08          & 0.587          & 0.336          &    \multicolumn{1}{c|}{1.711}          & 18.66          & 0.535          & 0.411                  & \multicolumn{1}{l}{1.998}        \\ \hline
FreeNeRF~\cite{yang2023freenerf}  
& 19.63          & 0.612          & 0.308          &  \multicolumn{1}{c|}{1.702}           & \cellcolor{top3}19.13          & 0.562          & \cellcolor{top3}0.384                & \multicolumn{1}{l}{1.914}        \\
SimpleNeRF~\cite{somraj2023simplenerf}   
& 19.24      &  0.623          & 0.375          &  \multicolumn{1}{c|}{-}            & -       & -      & -             & \multicolumn{1}{l}{-}       \\ 
SparseNeRF~\cite{wang2023sparsenerf}   
&19.86          &  0.624          & 0.328          &  \multicolumn{1}{c|}{1.628}            & 19.07          & \cellcolor{top3}0.564          & 0.392              & \multicolumn{1}{l}{\cellcolor{top3}1.901}      \\ 
ReconFusion~\cite{wu2024reconfusion}   
 &21.34          &  0.724          & 0.203         &  \multicolumn{1}{c|}{-}            & -       & -      & -             & \multicolumn{1}{l}{-}      \\ \hline
3D-GS~\cite{kerbl20233dggs}      
& 17.83          & 0.582          &     0.321   & \multicolumn{1}{c|}{\cellcolor{top2}1.481}            & 16.94          & 0.488          & 0.402                & \multicolumn{1}{l}{1.972}        \\
DNGaussian~\cite{li2024dngaussian}  
& 19.12          & 0.591          & 0.294          &  \multicolumn{1}{c|}{\cellcolor{top3}1.524}      & {- } & -         & -         &\multicolumn{1}{l}{ - }       \\  
FSGS~\cite{zhu2023fsgs}   
&  \cellcolor{top3}20.43          & \cellcolor{top3}0.682          &  \cellcolor{top3}0.248          & \multicolumn{1}{c|}{{1.571}}           &  \cellcolor{top2}19.71          &  \cellcolor{top2}0.642          &  \cellcolor{top2}0.283                    & \multicolumn{1}{l}{ \cellcolor{top2}1.872}        \\     
CoR-GS~\cite{zhang2025cor}   
&  \cellcolor{top2}20.45          & \cellcolor{top2}0.712          &  \cellcolor{top2}0.196          & \multicolumn{1}{c|}{-}           &  -          &  -          &  -                   & \multicolumn{1}{l}{ -}        \\        \hline
\textbf{Ours}                                & \textbf{\cellcolor{top1}22.01} & \textbf{\cellcolor{top1}0.728} & \textbf{\cellcolor{top1}0.175} &\multicolumn{1}{c|}{{\textbf{\cellcolor{top1}1.470}}} & \textbf{\cellcolor{top1}21.53} & \textbf{\cellcolor{top1}0.674} & \textbf{\cellcolor{top1}0.225}  & \multicolumn{1}{l}{\textbf{\cellcolor{top1}1.733}}   \\ \hline
\end{tabular}}
\label{NERF-LLFF}
\end{table*}

Besides, we also conducted a detailed qualitative analysis in Fig.~\ref{fig:teaser} and \ref{viz_360}.
3D-GS fails to obtain effective point clouds from sparse views, resulting in blurry rendering areas far from the camera (e.g. views outside the window).
 Although SparseNeRF~\cite{wang2023sparsenerf} is designed for sparse-view settings, it still struggles to reconstruct complex and fine-grained texture details of the stump. 
 FSGS~\cite{zhu2023fsgs}, designed global-local depth normalization for sparse-view, improves results but struggles with complex geometries and fine-grained texture.
 In contrast, our model outperforms in capturing complex structures of window views and fine textures of the stump, whether in indoor or outdoor scenes. 
 Besides, as shown in Fig.~\ref{more_depth}, we also present additional qualitative results of rendered depth on the Mip-NeRF360 dataset\cite{barron2022mip260}, comparing RDG-GS, 3D-GS~\cite{kerbl20233dggs}, FSGS~\cite{zhu2023fsgs}, and CoR-GS~\cite{zhang2025cor}. 
 Our approach consistently achieves more accurate geometric reconstructions and maintains robust view-consistency, attesting to its effectiveness in generating realistic scene geometry and ensuring coherent depth estimations across diverse viewpoints.

\noindent\textbf{Different Training Views.} Our comparative analysis across varying sparse training views, as delineated in Table~\ref{training_views_mip}, reveals that our approach achieves superior performance in PSNR, SSIM, and LPIPS metrics with 12 and 24 training views. Qualitative visualizations are presented in Fig.~\ref{views_viz_mip}. Notably, our model consistently attains optimal rendering outcomes under diverse sparse training view inputs, even in high-frequency areas with intricate texture details.

\begin{table*}[!htbp]
	\centering
 \caption{Evaluations of different 3, 6, 9 training views on NeRF-LLFF~\cite{mildenhall2019localllff} and DTU~\cite{6909453DTU} datasets.} 
      \setlength{\tabcolsep}{6pt}
	\scalebox{0.85}{
\begin{tabular}{cllllllllll}
\hline
\multicolumn{1}{l}{}  & \multicolumn{1}{c}{\multirow{2}{*}{Method}} & \multicolumn{3}{c|}{PSNR↑}                        & \multicolumn{3}{c|}{SSIM↑}                        & \multicolumn{3}{c}{LPIPS↓}                       \\
\multicolumn{1}{l}{}  & \multicolumn{1}{c}{}                        & 3-view         & 6-view         & \multicolumn{1}{c|}{9-view}         & 3-view         & 6-view         & \multicolumn{1}{c|}{9-view}         & 3-view         & 6-view         & 9-view         \\ \hline
\multirow{9}{*}{LLFF~\cite{mildenhall2019localllff}} & RegNeRF~\cite{niemeyer2022regnerf}                                     & 19.08          & 23.09          & \multicolumn{1}{c|}{24.84}          & 0.587          & 0.760           & \multicolumn{1}{c|}{0.820}           & 0.374          & 0.243          & 0.196          \\
                      & DiffusioNeRF~\cite{wynn2023diffusionerf}                                & 20.13          & 23.60           & \multicolumn{1}{c|}{24.62}          & 0.631          & 0.775          & \multicolumn{1}{c|}{0.807}          & 0.344          & 0.235          & 0.216          \\
                      & FreeNeRF~\cite{yang2023freenerf}                                    & 19.63          & 23.72          & \multicolumn{1}{c|}{25.12}          & 0.613          & 0.773          & \multicolumn{1}{c|}{0.820}           & 0.347          & 0.232          & 0.193          \\
                      & SimpleNeRF~\cite{somraj2023simplenerf}                                  & 19.24          & 23.05          & \multicolumn{1}{c|}{23.98}          & 0.623          & 0.737          & \multicolumn{1}{c|}{0.762}          & 0.375          & 0.296          & 0.286          \\
                      & ReconFusion~\cite{wu2024reconfusion}                                 & \cellcolor{top2}21.34          & \cellcolor{top3}24.25          & \multicolumn{1}{c|}{25.21}          & \cellcolor{top2}0.724          & 0.815          & \multicolumn{1}{c|}{0.848}          & \cellcolor{top3}0.203          & 0.152          & 0.134          \\ 
                      & 3DGS~\cite{kerbl20233dggs}                                        & 19.22          & 23.80           & \multicolumn{1}{c|}{\cellcolor{top3}25.44}         & 0.649          & 0.814          & \multicolumn{1}{c|}{\cellcolor{top3}0.860}           & 0.229          & \cellcolor{top3}0.125          & \cellcolor{top3}0.096          \\
                      & FSGS~\cite{zhu2023fsgs}                                       & 20.43          & 24.09          & \multicolumn{1}{c|}{25.31}          & 0.682          & \cellcolor{top3}0.823          & \multicolumn{1}{c|}{\cellcolor{top3}0.860}           & 0.248          & 0.145          & 0.122          \\
                      & CoR-GS~\cite{zhang2025cor}                                      & \cellcolor{top3}20.45          & \cellcolor{top2}24.49          & \multicolumn{1}{c|}{\cellcolor{top2}26.06}          & \cellcolor{top3}0.712          & \cellcolor{top2}0.837          & \multicolumn{1}{c|}{\cellcolor{top2}0.874}          & \cellcolor{top2}0.196          & \cellcolor{top2}0.115          & \cellcolor{top2}0.089          \\ 
                      & \textbf{Ours}                               & \textbf{\cellcolor{top1}22.01} & \textbf{\cellcolor{top1}26.72} & \multicolumn{1}{c|}{\textbf{\cellcolor{top1}28.01}} & \textbf{\cellcolor{top1}0.728} & \textbf{\cellcolor{top1}0.854} & \multicolumn{1}{c|}{\textbf{\cellcolor{top1}0.883}} & \textbf{\cellcolor{top1}0.175} & \textbf{\cellcolor{top1}0.107} & \textbf{\cellcolor{top1}0.073} \\ \hline
\multirow{8}{*}{DTU~\cite{6909453DTU}}  & RegNeRF~\cite{niemeyer2022regnerf}                                     & 19.39          & 22.24          & \multicolumn{1}{c|}{24.62}          & 0.777          & 0.850           & \multicolumn{1}{c|}{0.886}          & 0.203          & 0.135          & 0.106          \\
                      & DiffusioNeRF~\cite{wynn2023diffusionerf}                                & 16.14          & 20.12          & \multicolumn{1}{c|}{24.31}          & 0.731          & 0.834          & \multicolumn{1}{c|}{0.888}          & 0.221          & 0.150           & 0.111          \\
                      & FreeNeRF~\cite{yang2023freenerf}                                    & \cellcolor{top3}20.46          & 23.48          & \multicolumn{1}{c|}{25.56}          & 0.826          & 0.870           & \multicolumn{1}{c|}{0.902}          & 0.173          & 0.131          & 0.102          \\
                      & SimpleNeRF~\cite{somraj2023simplenerf}                                  & 16.25          & 20.60           & \multicolumn{1}{c|}{22.75}          & 0.751          & 0.828          & \multicolumn{1}{c|}{0.856}          & 0.249          & 0.190           & 0.176          \\
                      & ReconFusion~\cite{wu2024reconfusion}                                 & \cellcolor{top2}20.74          & 23.61          & \multicolumn{1}{c|}{24.62}          & \cellcolor{top2}0.875          & 0.904          & \multicolumn{1}{c|}{0.921}          & \cellcolor{top3}0.124          & 0.105          & 0.094          \\ 
                      & 3DGS~\cite{kerbl20233dggs}                                        & 17.65          & \cellcolor{top3}24.00             & \multicolumn{1}{c|}{\cellcolor{top3}26.85}          & 0.816          & \cellcolor{top3}0.907          & \multicolumn{1}{c|}{\cellcolor{top3}0.942}          & 0.146          & \cellcolor{top3}0.076          & \cellcolor{top3}0.049          \\
                      & CoR-GS~\cite{zhang2025cor}                                      & 19.21          & \cellcolor{top2}24.51          & \multicolumn{1}{c|}{\cellcolor{top2}27.18}          & \cellcolor{top3}0.853          & \textbf{\cellcolor{top1}0.917}          &\multicolumn{1}{c|}{\cellcolor{top2} 0.947}          & \cellcolor{top2}0.119          & \cellcolor{top2}0.068          & \cellcolor{top2}0.045          \\ 
                      & \textbf{Ours}                               & \textbf{\cellcolor{top1}23.51} & \textbf{\cellcolor{top1}25.73} & \multicolumn{1}{c|}{\textbf{\cellcolor{top1}28.32}} & \textbf{\cellcolor{top1}0.881} & {\cellcolor{top2}0.914} & \multicolumn{1}{c|}{\textbf{\cellcolor{top1}0.951}} & \textbf{\cellcolor{top1}0.113} & \textbf{\cellcolor{top1}0.059} & \textbf{\cellcolor{top1}0.039} \\ \hline
\end{tabular}}
	\label{training_views}
\end{table*}

\begin{table}[!htbp]
 \caption{Comparison with geometry-aware 3DGS methods to sparse-view setups on Mip-NeRF~\cite{somraj2023simplenerf} and LLFF~\cite{mildenhall2019localllff} datasets.} 
      \setlength{\tabcolsep}{10pt}
	\setlength{\tabcolsep}{0.5mm}{
\begin{tabular}{lcccccc}
\hline
\multicolumn{1}{c}{}                         & \multicolumn{3}{c|}{Mip-NeRF360} & \multicolumn{3}{c}{NeRF-LLFF} \\ 
\multicolumn{1}{c}{\multirow{-2}{*}{Method}} & PSNR↑    & SSIM↑    & \multicolumn{1}{c|}{LPIPS↓}    & PSNR↑    & SSIM↑   & LPIPS↓   \\ \hline
SuGaR~\cite{guedon2024sugar}                                        & 19.43    & 0.619    & \multicolumn{1}{c|}{0.376}     & 16.33    & 0.475   & 0.423    \\
{2DGS~\cite{huang20242d}}                  & 20.58    & 0.647    & \multicolumn{1}{c|}{0.311}     & 17.18    & 0.492   & 0.388    \\
{ GOF~\cite{yu2024gaussiangof}}                   & \cellcolor{top2}22.16    & \cellcolor{top3}0.702    & \multicolumn{1}{c|}{\cellcolor{top3}0.258}     &  \cellcolor{top2}18.64    &  \cellcolor{top2}0.553   & \cellcolor{top3}0.332    \\
{PGSR~\cite{chen2024pgsr}}                  & \cellcolor{top3}22.01    &  \cellcolor{top2}0.711    & \multicolumn{1}{c|}{{\cellcolor{top2}0.249}}     &\cellcolor{top3} 18.29    & \cellcolor{top3}0.541   &  \cellcolor{top2}0.298    \\ \hline
\textbf{Ours}                                         & \textbf{\cellcolor{top1}26.03}    & \textbf{\cellcolor{top1}0.794}    & \multicolumn{1}{c|}{\textbf{\cellcolor{top1}0.219} }    & \textbf{\cellcolor{top1}21.53}    & \textbf{\cellcolor{top1}0.674}   & \textbf{\cellcolor{top1}0.225}    \\ \hline
\end{tabular}
}
	\label{gemtorty methods}
\end{table}

\noindent \textbf{Geometry and Fine Textures.}
Besides, we will provide a qualitative visualization to highlight our model's superiority in capturing superior geometry and fine textures in Fig.~\ref{more_depth}. 
As depicted in Fig.~\ref{more_depth} (a) and (b), we demonstrate the superiority of our proposed RDG-GS in generating accurate geometry, as well as its capability to produce superior fine fine-grained details for intricate and complex scenes in Fig.~\ref{more_depth} (c) and (d). While 3D Gaussian Splatting~\cite{kerbl20233dggs} exhibits partial inaccuracies in generating these complex scenes, FSGS~\cite{zhu2023fsgs} and CoR-GS~\cite{zhang2025cor} fail to capture high-frequency detail information. This comparison further illustrates our model's ability to generate high-quality scenes with correct geometric shapes and fine details.

\subsubsection{Comparison on DTU}
Table~\ref{scannet and bender} illustrates that our model quantitatively surpasses all state-of-the-art approaches~\cite{kerbl20233dggs,li2024dngaussian,zhu2023fsgs,zhang2025cor,yang2023freenerf,wu2024reconfusion,wang2023sparsenerf} on the DTU dataset. Notably, compared to FSGS~\cite{zhu2023fsgs} and DNGaussian~\cite{li2024dngaussian}, which also leverage depth regularization, our model achieves a significant performance advantage, attributed to the refined depth optimization and relative depth guidance.

\begin{figure}[!htbp]
\centering
\includegraphics[width=1\linewidth]{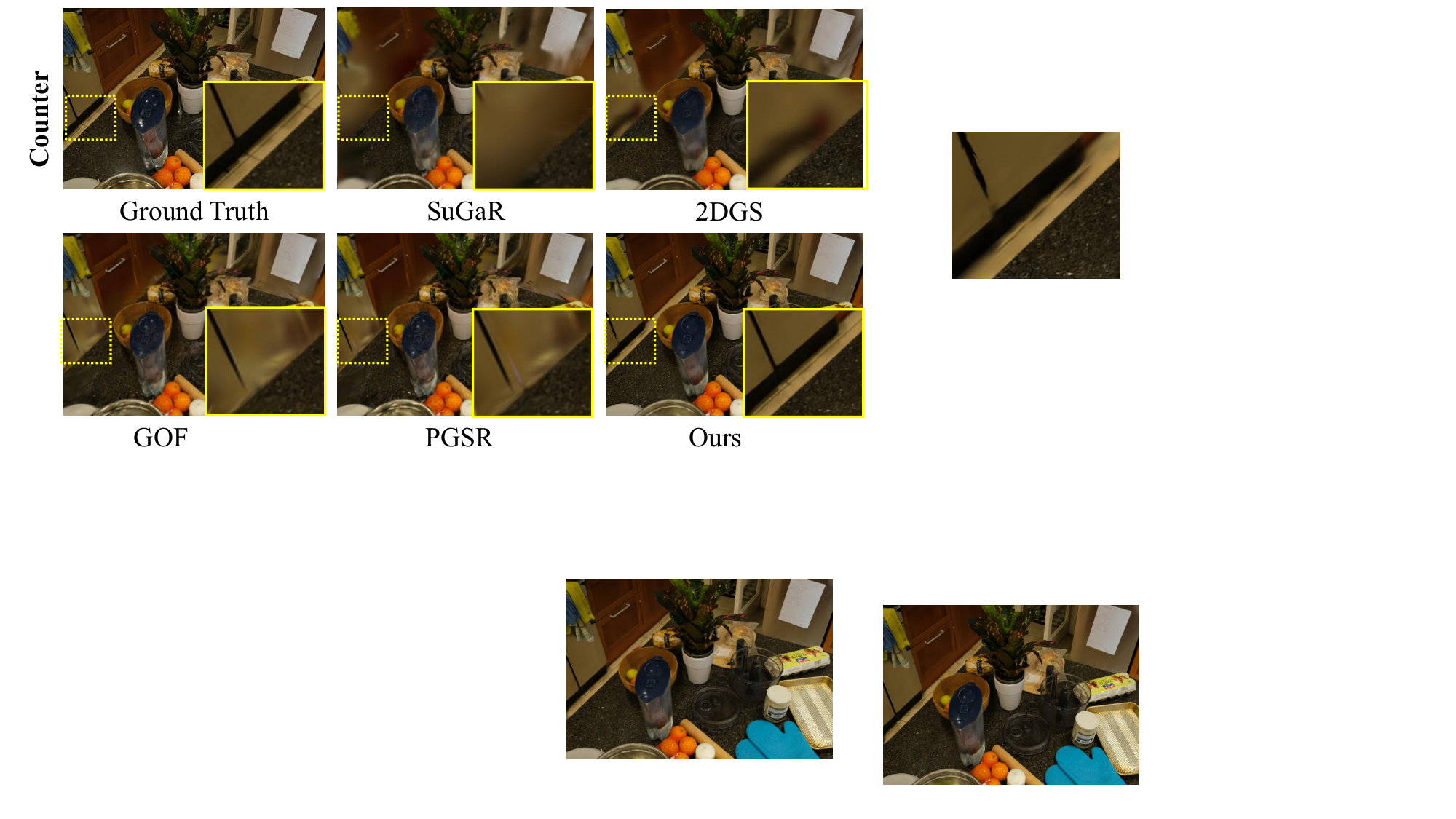}
\caption{Qualitative comparison with geometry-aware 3DGS
methods~\cite{guedon2024sugar,huang20242d,yu2024gaussiangof,chen2024pgsr} to sparse-view setups on Mip-NeRF dataset~\cite{somraj2023simplenerf}. }
\label{geo_method}
\end{figure}
We also present detailed qualitative results in Fig.~\ref{viz_blender} (A), which demonstrate that 3D-GS~\cite{kerbl20233dggs}, FSGS~\cite{zhu2023fsgs}, and the SOTA NeRF-based SparseNeRF~\cite{wang2023sparsenerf}, struggle to accurately capture detailed fine-grained information for the objects in whole scenes, such as bottle caps and balls. In contrast, our model exhibits significant advantages in recovering fine geometric shapes, such as the rims of bottle caps, the orange, and balls.

\begin{table*}[h]
	\centering
 \caption{Comparison of efficiency and costs on Mip-NeRF360~\cite{barron2022mip260} dataset. The term ``Backbone" refers to classical pipeline methods such as 3D-GS~\cite{kerbl20233dggs}. In our work, we employ distinct methods to demonstrate that the effectiveness of our approach stems from the proposed strategy rather than the backbone itself, enabling a more rigorous comparison.}  
      \setlength{\tabcolsep}{10pt}
	\scalebox{0.95}{
\begin{tabular}{lcccccc}
\hline
\multicolumn{1}{c}{\multirow{2}{*}{Backbone}} & \multirow{2}{*}{Methods} & \multicolumn{5}{c}{Mip-NeRF360:1/2× Resolution}           \\
\multicolumn{1}{c}{}                          &                          & PSNR↑ & SSIM↑ & LPIPS↓ & FPS↑ & Costs (GB)↓ \\ \hline
FreeNeRF~\cite{yang2023freenerf}                                 & /                        & 18.35 & 0.476 & 0.514  & 0.03       & 192        \\
SparseNeRF~\cite{wang2023sparsenerf}                                    & /                        & 19.02 & 0.497 & 0.476  & 0.03        & 192        \\ \hline
\multirow{5}{*}{3D-GS~\cite{kerbl20233dggs}}                        & None                     & 18.35 & 0.476 & 0.514  & \cellcolor{top3}122         & 8          \\
                                              & FreeNeRF~\cite{yang2023freenerf}                 & 18.67 & 0.484 & 0.502  & 92          & 8          \\
                                              & SparseNeRF~\cite{wang2023sparsenerf}               & 18.82 & \cellcolor{top3}0.497 & \cellcolor{top3}0.489  & 84     & 8          \\
                                              & FSGS~\cite{zhu2023fsgs}                 & \cellcolor{top2}20.11 & \cellcolor{top2}0.511 & \cellcolor{top2}0.414  & \textbf{\cellcolor{top1}148}         & 8          \\
                                              & DNGaussian~\cite{li2024dngaussian}               & \cellcolor{top3}18.83 & 0.484 & 0.502  & \cellcolor{top2}124        & \textbf{4}          \\ \hline
                                         \textbf{3DGS~\cite{kerbl20233dggs}}                                 & \textbf{Ours}            & {\textbf{\cellcolor{top1}22.67}} & \textbf{\cellcolor{top1}0.548} & \textbf{\cellcolor{top1}0.354} & {112}  &{8} \\ \hline
\end{tabular}}

	\label{work_transfer}
\end{table*}

\begin{table*}[!h]
 \caption{Comparison of training computation time of per scene on NeRF-LLFF~\cite{mildenhall2019localllff}, DTU~\cite{6909453DTU}, and Blender~\cite{zhang2018unreasonablelpips} datasets.} 
	\small
    \setlength{\tabcolsep}{4mm}{
\begin{tabular}{lcccccc}
\hline
\multicolumn{1}{c}{\multirow{2}{*}{Method}} & \multicolumn{2}{c}{NeRF-LLFF} & \multicolumn{2}{c}{DTU}       & \multicolumn{2}{c}{Blender}   \\  
\multicolumn{1}{c}{}                        & PSNR           & Time (min)↓  & PSNR           & Time (min)↓  & PSNR           & Time (min)↓  \\ \hline
RegNeRF~\cite{niemeyer2022regnerf}                                     & 19.08          & 118          & 18.89          & 139          & 23.86          & 128          \\
FreeNeRF~\cite{yang2023freenerf}                                    & 19.63          & 133          & \cellcolor{top3}19.92          & 162          & 24.26          & 148          \\
SparseNeRF~\cite{wang2023sparsenerf}                                  & \cellcolor{top3}19.86          & 105          & 19.55          & 134          & 24.04          & 136          \\ \hline
3D-GS~\cite{kerbl20233dggs}                                       & 17.83          & \textbf{\cellcolor{top1}2.7} & 10.99          & \textbf{\cellcolor{top1}4.5} & 21.56          & \textbf{\cellcolor{top1}5.3} \\
DNGaussian~\cite{li2024dngaussian}                                  & 19.12          & 3.5          & 18.91          & 5.5          & \cellcolor{top3}24.31          & 6.5          \\
FSGS~\cite{zhu2023fsgs}                                        & \cellcolor{top2}20.43          & 4.5          & \cellcolor{top2}21.21          & 7.8          & \cellcolor{top2}24.64          & 8.7          \\ \hline
Ours                                        & \textbf{\cellcolor{top1}22.01} & 4.9          & \textbf{\cellcolor{top1}23.51} & 8.3          & \textbf{\cellcolor{top1}26.21} & 9.8          \\ \hline
\end{tabular}
}
	\label{training time}
\end{table*}

\subsubsection{Comparison on Blender}
From Table~\ref{scannet and bender}, it can be observed that our model outperforms other SOTA models~\cite{niemeyer2022regnerf,wu2024reconfusion,zhang2025cor,zhu2023fsgs,li2024dngaussian} on the Blender dataset~\cite{mildenhall2019localllff}, demonstrating its superior ability to generate geometric scenes with photorealistic objects. 
Compared to MVSNeRF~\cite{chen2021mvsnerf} designed for large-scale scenes with complex geometry, we achieve a $1.88$ PSNR improvement. We also provide a qualitative comparison in Fig.~\ref{viz_blender} (B), which clearly illustrates that 3D-GS~\cite{kerbl20233dggs} fails to recognize the complex geometric shape of legos, while SparseNeRF~\cite{wang2023sparsenerf} exhibits many blurry areas of wheels. FSGS~\cite{zhu2023fsgs} is unable to render shadow/light effects, with uncertain floaters.  In contrast, our model not only generates view-consistent geometric information for complex objects (e.g. lego and Hotdog) but also restores fine-grained details (e.g. wheels).

\subsubsection{Comparison on NeRF-LLFF} We conduct comparisons on NeRF-LLFF~\cite{mildenhall2019localllff} with $3$ training views in Table~\ref{NERF-LLFF}. It can be seen that our model achieves significant improvements across different resolutions. In cases of $3$ input views of $503 \times 381$ resolution, our model exhibits a $1.56$ PSNR improvement compared to the SOTA 3D-GS based work CoR-GS~\cite{zhang2025cor}. 

We provide a detailed quantitative analysis in Fig.~\ref{viz_llff}. The results demonstrate that 3D-GS\cite{kerbl20233dggs} struggles to accurately reconstruct structures, resulting in geometric inaccuracies, particularly in the depiction of leaf shapes. FSGS~\cite{zhu2023fsgs} is designed for sparse views, which still struggles to capture the fine-grained texture of leaves and the reflection of flowers. In comparison, our model can reconstruct high-quality scenes and credible geometric shapes, even in complex and shadow areas.

\noindent \textbf{Training Views.}
We further investigated the performance of the model under different input views, as shown in Table~\ref{training_views}. In addition to the 3 views utilized in the paper, we conducted experiments on the LLFF dataset~\cite{mildenhall2019localllff} with 6 and 9 input views. Detailed comparative experiments reveal that as the number of views increases, the performance of all the models improves. 
Notably, our model consistently outperforms others~\cite{kerbl20233dggs,zhu2023fsgs,zhang2025cor,somraj2023simplenerf,wu2024reconfusion,yang2023freenerf} across different sparse view settings, demonstrating its capability to effectively reconstruct optimal scenes from limited input.

 \begin{figure*}[!htbp]
\centering
\includegraphics[width=1\linewidth]{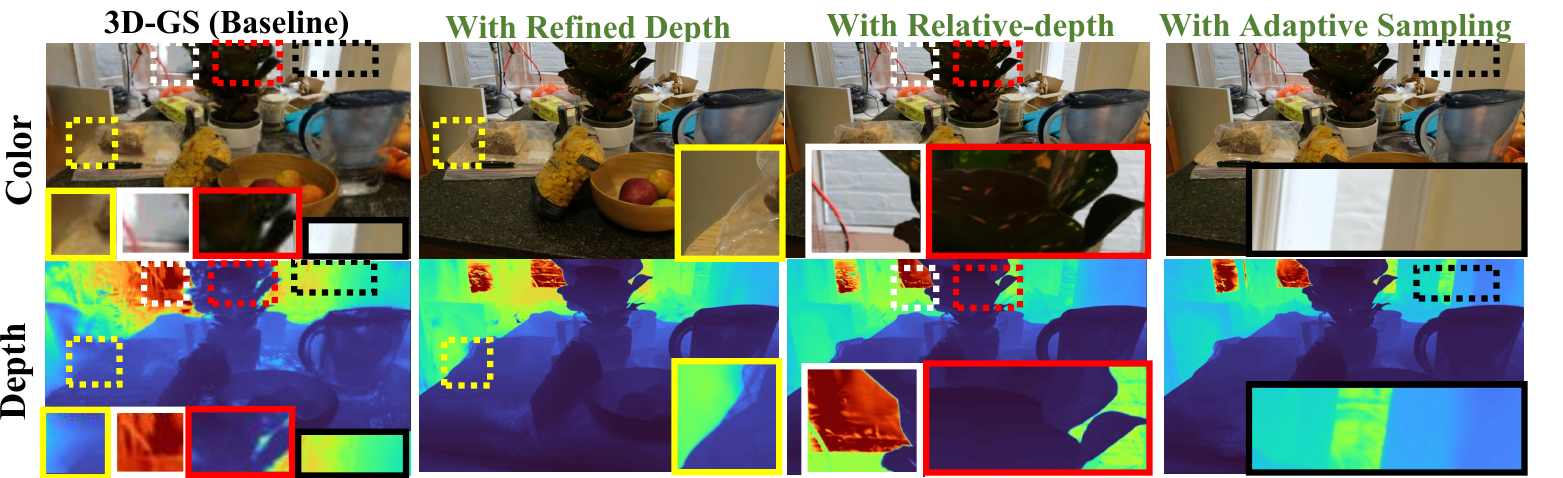}
\caption{
Ablation Study of Quantitative Comparison. Using 3D-GS~\cite{kerbl20233dggs} as the baseline, we present the performance of the baseline model integrated with each of our proposed modules, which namely, refined depth regulation, relative refined depth guidance, and adaptive sampling, on the Mip-NeRF dataset~\cite{barron2022mip260}.}
\label{viz_ab}
\end{figure*}

\begin{table*}
\centering
 \caption{Ablation study. ``Coarse": Coarse depth, ``Refined": Refined depth, ``RDG”: Relative Depth Guidance, ``AS”: Adaptive Sampling. } 
	\scalebox{0.95}{
	\setlength{\tabcolsep}{8pt}
\begin{tabular}{cccccccccc}
\hline
\multicolumn{2}{c}{Depth} & \multirow{2}{*}{RDG} & \multirow{2}{*}{AS} & \multicolumn{3}{c|}{Mip-NeRF360: 1/8 Resolution}                               & \multicolumn{3}{c}{NeRF-LLFF: 1006 × 762}                                        \\ 
Coarse      & Refined     &                      &                      & PSNR↑   &  SSIM↑     & \multicolumn{1}{c|}{ LPIPS↓}                       & PSNR↑          & SSIM↑          & LPIPS↓                  \\ \hline
×           & ×           & ×                    & ×                    & 20.89     & 0.633    & \multicolumn{1}{c|}{ 0.317}                      & 16.94          & 0.488          & 0.402                       \\ \hline
\checkmark           & ×           & ×                    & ×                    & 21.43  &  0.654      &  \multicolumn{1}{c|}{0.311}                         & 17.33          & 0.503          & 0.388                    \\
\checkmark           & ×           & \checkmark                     & ×                    & 23.15  &  0.679     &  \multicolumn{1}{c|}{0.385}                         &   18.47     &  0.546     &    0.370                \\
\checkmark           & ×           & ×                  &  \checkmark                    & 23.52  &  0.698     &  \multicolumn{1}{c|}{0.378}                         & 19.05       &   0.562    &    0.359                \\ 
\checkmark           & ×           & \checkmark                     & \checkmark                       & 24.38  & 0.725      &  \multicolumn{1}{c|}{0.352}                         &  19.58      &  0.599     &       0.322             \\  \hline

×           & \checkmark           & ×                    & ×                    & 22.68  &  0.696      &  \multicolumn{1}{c|}{0.271}                        & 18.67          & 0.548          & 0.302                      \\
×           & \checkmark           & \checkmark                    & ×                    & 24.81      & 0.731   &  \multicolumn{1}{c|}{0.233}                   &20.22          & 0.614          & 0.269                   \\
×           & ×           & ×                    & \checkmark                    & 22.55      &  0.671  &  \multicolumn{1}{c|}{0.288}             & 19.01          & 0.532          & 0.349                    \\
×           & \checkmark           & ×                    & \checkmark                    & 25.11    &  0.753   & \multicolumn{1}{c|}{0.251}                 & 20.94             & 0.629               &0.271       \\
\textbf{×}  & \textbf{\checkmark}  & \textbf{\checkmark}           & \textbf{\checkmark}           & \textbf{26.03}&\textbf{0.794} &  \multicolumn{1}{c|}{\textbf{0.219}} & \textbf{21.53} & \textbf{0.674} & \textbf{0.225} \\ \hline
\end{tabular}}
\label{ablation}
\end{table*}

\begin{figure}[!htbp]
\centering
\includegraphics[width=1\linewidth]{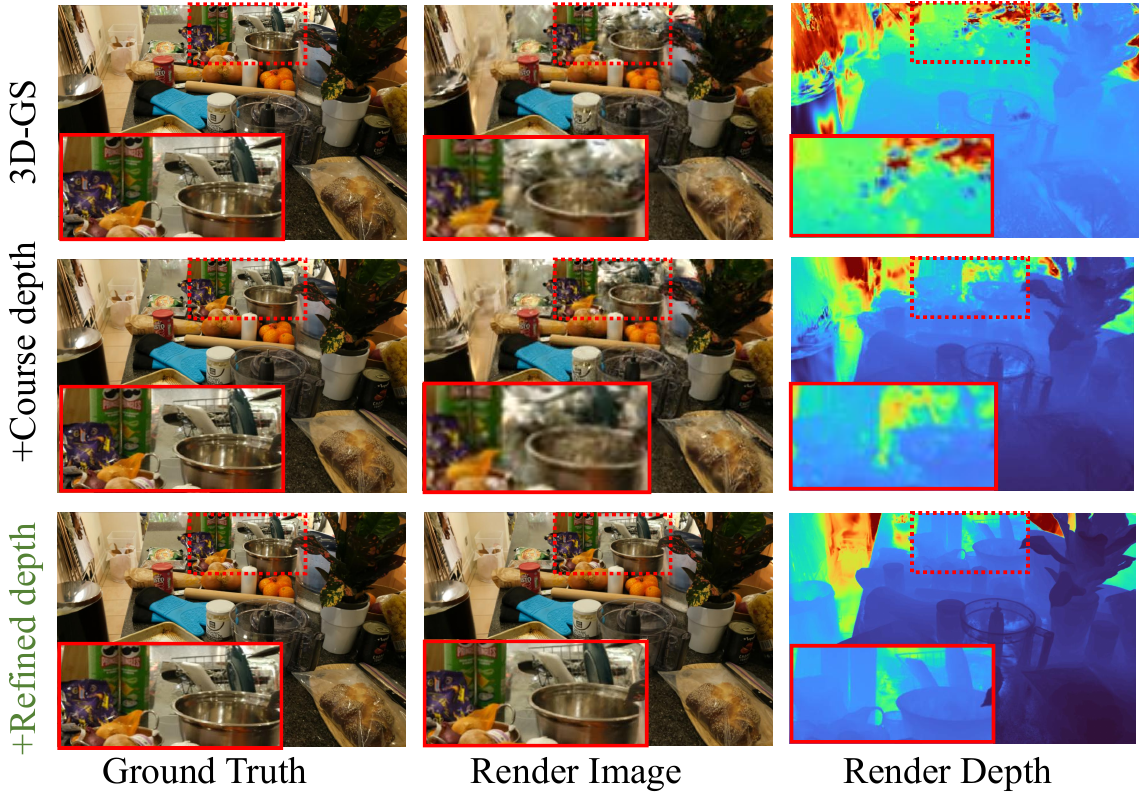}
\caption{Comparison of coarse monocular estimated depth employed by general SOTA models~\cite{zhu2023fsgs,li2024dngaussian,guo2024depth} and our refined depth.  }
\label{c_depth}
\end{figure}

\subsubsection{Comparison with Geometry-aware Methods}
To ensure a fair comparison, we extend 4 robust geometry-aware methods~\cite{guedon2024sugar,huang20242d,yu2024gaussiangof,chen2024pgsr} to sparse-view reconstruction and present a detailed comparison under identical settings. Experiments were conducted on the Mip-NeRF~\cite{barron2022mip260} with 24 training views and NeRF-LLFF with 3 training views.
As illustrated in Table~\ref{gemtorty methods}, our model demonstrates a clear performance advantage over state-of-the-art (SOTA) geometry-aware 3D Gaussian Splatting (3DGS) methods when evaluated under identical sparse-view configurations. Notably, it achieves a significant PSNR improvement of 3.24 over PGSR~\cite{chen2024pgsr} on the  LLFF dataset~\cite{mildenhall2019localllff}.  
By leveraging enhanced geometric learning, our model excels in capturing view-consistent geometry, which is critical for accurate and reliable 3D reconstruction from limited input views.

Moreover, we evaluated the geometric reconstruction performance of these models on the complex unbounded scenes in the Mip-NeRF dataset. As illustrated in Fig.~\ref{gemtorty methods}, the geometry-aware 3DGS methods~\cite{guedon2024sugar,huang20242d,yu2024gaussiangof,chen2024pgsr} struggle to extract sufficient geometric information from sparse-view inputs, resulting in blurred and malformed geometries. In contrast, our model successfully renders view-consistent geometric structures, attributed to the proposed refined depth and relative depth guidance mechanisms, which effectively capture comprehensive and accurate global geometric information.

\begin{figure*}[!htbp]
\centering
\includegraphics[width=1\linewidth]{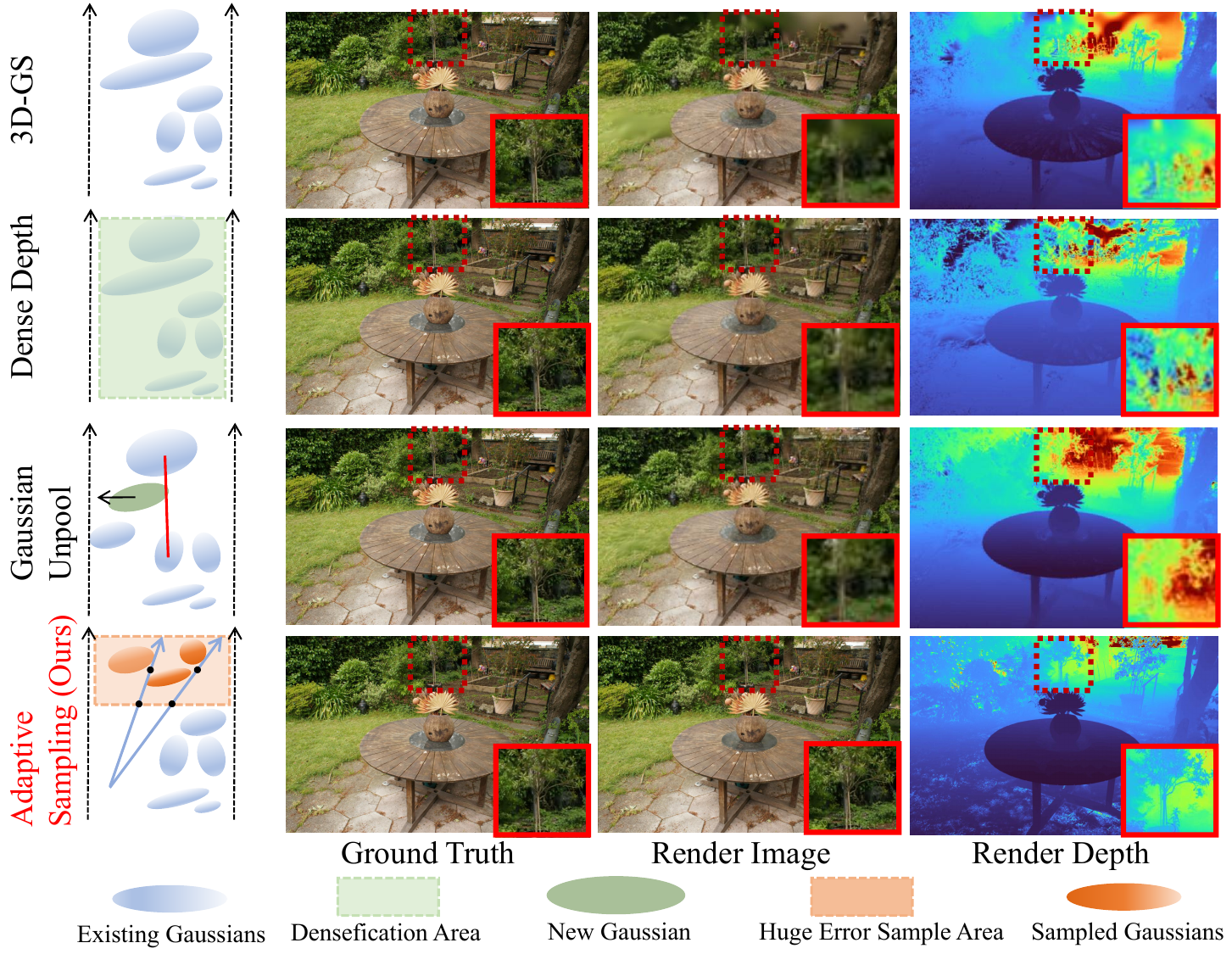}
    \caption{Comparison of density initialization methods, encompassing dense initialization with COLMAP~\cite{kerbl20233dggs}, Gaussian Unpooling initialization from FSGS~\cite{zhu2023fsgs}, and the proposed adaptive sampling approach. }
\label{Comparision of the initialization}
\end{figure*}

\subsubsection{Efficiency and Costs}
We also evaluated efficiency, as shown in Table \ref{work_transfer} and \ref{training time}. Compared to SOTA NeRF-based methods~\cite{yang2023freenerf,wang2023sparsenerf}, we achieve over $3500\times$ acceleration in FPS. While our FPS is slightly lower than other GS-based methods~\cite{zhu2023fsgs,li2024dngaussian}, we deliver superior rendering of high quality with greater cost-efficiency.

We also compare the average training time per scene between our model and SOTA NeRF-based~\cite{niemeyer2022regnerf,yang2023freenerf,wang2023sparsenerf} and GS-based~\cite{kerbl20233dggs,li2024dngaussian,zhu2023fsgs} models under the same sparse-view settings. As shown in Table~\ref{training time}, our method achieves a training speed nearly 20-40 times faster than NeRF-based approaches~\cite{niemeyer2022regnerf,yang2023freenerf,wang2023sparsenerf}, reducing training time from hours to mere minutes. Although our method takes approximately 2 times longer than the original 3D-GS~\cite{kerbl20233dggs}, it remains cost-effective while delivering superior rendering quality. Future work will focus on further optimizing training efficiency.

\subsection{Ablation Study}

As Table~\ref{ablation} and Fig.~\ref{viz_ab} show,  we conduct ablation studies to validate the efficacy of the proposed refined depth, relative depth guidance, and the adaptive sampling strategy. We take the 3D-GS~\cite{kerbl20233dggs} as the baseline, as shown in row $1$.

\begin{table*}[h]
	\centering
 \caption{
 Comparisons of different types of depth monocular models. We adopt DPT~\cite{ranftl2021visiondpt}, a depth estimation model widely employed by state-of-the-art NeRF-based~\cite{wang2023sparsenerf} and GS-based works~\cite{zhu2023fsgs,li2024dngaussian} To further evaluate the robustness of our proposed refined depth, we also performed ablation experiments using various DPT models, including \emph{dpt-hybrid-384} and \emph{dpt-large -384}.}  
      \setlength{\tabcolsep}{4pt}
	\scalebox{1}{
\begin{tabular}{llccc|ccc}
\hline
\multicolumn{2}{c}{\multirow{2}{*}{Methods}} & \multicolumn{3}{c|}{Mip-NeRF360~\cite{barron2022mip260}:1/2 Resolution}  & \multicolumn{3}{c}{LLFF~\cite{mildenhall2019localllff}:1006 × 762 Resolution}   \\  
\multicolumn{2}{c}{}                         & PSNR↑          & SSIM↑          & LPIPS↓         & PSNR↑          & SSIM↑          & LPIPS↓         \\ \hline
\multirow{2}{*}{FSGS~\cite{zhu2023fsgs}}          & \emph{dpt-hybrid-384}      & 19.56          & 0.492          & 0.432          & 19.71          & 0.642          & 0.283          \\
                               & \emph{dpt-large-384}      &\cellcolor{top3} \cellcolor{top3}20.11          &\cellcolor{top3} 0.511          & \cellcolor{top3}0.414          & \cellcolor{top3}19.59          &\cellcolor{top3} 0.629          & \cellcolor{top3}0.271          \\ \hline
\multirow{2}{*}{Ours}       & \emph{dpt-hybrid-384}      & \textbf{\cellcolor{top2}22.62} & \textbf{\cellcolor{top2}0.542} & \textbf{\cellcolor{top2}0.321} & \textbf{\cellcolor{top2}21.53} & \textbf{\cellcolor{top2}0.674} & \textbf{\cellcolor{top2}0.225} \\
                               & \emph{dpt\_large-384}      & \textbf{\cellcolor{top1}22.67} & \textbf{\cellcolor{top1}0.548} & \textbf{\cellcolor{top1}0.354} & \textbf{\cellcolor{top1}21.51} & \textbf{\cellcolor{top1}0.670}  & \textbf{\cellcolor{top1}0.223} \\ \hline
\end{tabular}}
	\label{depth_model}
\end{table*}

 \begin{table*}
	\centering
  \small
 \caption{Influence of different initialization methods. Including the dense initialization based on COLMAP, Gaussian Unpooling initialization utilized by FSGS~\cite{zhu2023fsgs}, and our proposed adaptive sampling initialization method.} 
      \setlength{\tabcolsep}{6.5pt}
	\scalebox{1}{
\small
\begin{tabular}{lccc|ccc}
\hline
\multirow{2}{*}{Initialization Methods}    & \multicolumn{3}{c|}{Mip-NeRF360~\cite{barron2022mip260}:1/4 Resolution}  & \multicolumn{3}{c}{DTU~\cite{6909453DTU}}                          \\
                                           & PSNR↑          & SSIM↑          & LPIPS↓         & PSNR↑          & SSIM↑          & LPIPS↓         \\ \hline
Dense initialization                       & \cellcolor{top3}21.76         & \cellcolor{top3}0.648          & \cellcolor{top3}0.324          & \cellcolor{top3}20.43         & \cellcolor{top3}0.721          &\cellcolor{top3} 0.196          \\
Gaussian Pooling                      & \cellcolor{top2}22.52          &\cellcolor{top2} 0.673          & \cellcolor{top2}0.313          &\cellcolor{top2} 21.21          & \cellcolor{top2}0.782          & \cellcolor{top2}0.172          \\ \hline
Random Sampling                            & 19.15          & 0.603          & 0.352          & 16.82          & 0.643          & 0.235          \\
\textbf{Adaptive Sampling (Ours)} & \textbf{\cellcolor{top1}25.01} & \textbf{\cellcolor{top1}0.738} & \textbf{\cellcolor{top1}0.245} & \textbf{\cellcolor{top1}23.51} & \textbf{\cellcolor{top1}0.818} & \textbf{\cellcolor{top1}0.147} \\ \hline
\end{tabular}}
	\label{initialization}
\end{table*}

\subsubsection{Refined Depth}
From the comparison of rows 1 and 6, we can obverse that the baseline equipped with the refined RGB-guided depth regulation obtains the $1.79$ and $1.73$ PSNR improvements. 
We show the quantitative comparison in Fig.~\ref{viz_ab}, where the baseline equipped with our refined depth can accurately render the geometric information of the scene and the fine-grained shape of plastic bags (\textcolor{yellow}{yellow box}), highlighting the effectiveness of the depth correction.

\noindent\textbf{Depth Comparison.}
We also compared the general coarse depth supervision with our refined depth supervision, as shown in rows 2 and 6 in Table~\ref{ablation}, which shows a significant improvement. As illustrated in Fig.~\ref{c_depth}, we further discuss the reconstruction results obtained using our proposed refined depth against the coarse depth employed by general SOTA models~\cite{zhu2023fsgs,li2024dngaussian,guo2024depth}. We meticulously visualize both the rendered images and depths of the models, demonstrating that refined depth allows for the rapid acquisition of accurate geometric information and reconstruction of geometrically consistent scenes, whereas coarse depth suffers from estimation errors, leading Gaussian splats to render the incorrect geometric scenes.
Furthermore, we conducted experiments cooperating coarse depth with our proposed relative depth guidance and adaptive sampling strategies, as detailed in rows 2-5 of Table~\ref{ablation}.   The results indicate that while combining our RDG and AS strategies indeed provides some improvement, the coarse estimated depth still fails to effectively optimize the Gaussian field in terms of accurate geometry and fine details when compared to refined depth.

\subsubsection{Relative Depth Guidance}

In rows 6 and 7 of Table~\ref{ablation}, we observe that incorporating relative depth guidance strategy increases the PSNR values on Mip-NeRF360~\cite{barron2022mip260} and NeRF-LLFF~\cite{mildenhall2019localllff} datasets by 2.13 and 1.55, respectively. Furthermore, the visual comparisons in Fig.~\ref{viz_ab} confirm that this guidance strategy yields superior rendering quality, with accurately reconstructed geometric structures and faithful spatial relationships (as illustrated by the \textcolor{red}{red} and \textcolor{black}{white} boxes). Overall, these results underscore the potency of relative depth guidance for achieving view-consistent geometry and preserving precise spatial correlations.


\begin{table*}[!htp]
	\centering
 \caption{Evaluations of different spherical harmonic (SH) on NeRF-LLFF~\cite{mildenhall2019localllff} of 3 training views. We compare against the specialized neural color renderer in DNGaussian~\cite{li2024dngaussian} to substantiate our model’s superior reconstruction performance.} 
      \setlength{\tabcolsep}{19pt}
	\scalebox{1}{
\begin{tabular}{ccccc}
\hline
\multirow{2}{*}{Method}     & \multirow{2}{*}{Setting} & \multicolumn{3}{c}{LLFF~\cite{mildenhall2019localllff}: 503 × 381 Resolution}   \\
                            &                          & PSNR↑          & SSIM↑          & LPIPS↓         \\ \hline
\multirow{3}{*}{DNGaussian~\cite{li2024dngaussian}} & SH degree=2              & 17.06          & 0.549          & 0.333          \\
                            & SH degree=3              & 17.11          & 0.563          & 0.328          \\
                            & Neural Renderer          & 19.12          & 0.591          & 0.294          \\ \hline
\multirow{3}{*}{Ours}       & SH degree=2              & \cellcolor{top3}22.01          & \cellcolor{top3}0.728          &\cellcolor{top3} 0.175          \\
                            & SH degree=3              & \cellcolor{top2}22.03          &\cellcolor{top2} 0.734          & \cellcolor{top2}0.171          \\
                            & Neural Renderer          & \textbf{\cellcolor{top1}22.32} & \textbf{\cellcolor{top1}0.745} & \textbf{\cellcolor{top1}0.167} \\ \hline
\end{tabular}}
	\label{spherical harmonic}
\end{table*}

\begin{table*}[!h]
\centering
 \caption{Comparison with different hyperparameters of the weights $w_p$ and $w_h$ in Eqn.~\ref{energy} on Mip-NeRF~\cite{somraj2023simplenerf} and LLFF~\cite{mildenhall2019localllff} datasets.} 
      \setlength{\tabcolsep}{80pt}
	\setlength{\tabcolsep}{4.5mm}{
\begin{tabular}{cccccccc}
\hline
\small
\multirow{2}{*}{$w_p$}        & \multirow{2}{*}{$w_h$} & \multicolumn{3}{c}{Mip-NeRF360}                  & \multicolumn{3}{c}{NeRF-LLFF:}                   \\
                             &                       & PSNR↑          & SSIM↑          & LPIPS↓         & PSNR↑          & SSIM↑          & LPIPS↓         \\ \hline
\multirow{3}{*}{5}           & 5                     & 25.53          & 0.778          & 0.231          & 21.17          & 0.665          & 0.235          \\
                             & 10                    & 25.81          & 0.789          & 0.227          & 21.32          & 0.671          & 0.230          \\
                             & 15                    & 25.38          & 0.771          & 0.239          & 21.08          & 0.653          & 0.245          \\ \hline
\multirow{3}{*}{\textbf{10}} & \textbf{5}                     & \textbf{26.03} & \textbf{0.794} & \textbf{0.219} & \textbf{21.53} & \textbf{0.674} & \textbf{0.225}      \\
                             & {10}           & 25.78          & 0.785          & 0.223          & 21.33          & 0.668          & 0.231     \\
                             & 15                    & 25.52          & 0.776          & 0.231          & 21.29          & 0.659          & 0.239          \\ \hline
\multirow{3}{*}{15}          & 5                     & 21.61          & 0.775          & 0.238          & 21.06          & 0.662          & 0.241          \\
                             & 10                    & 25.89          & 0.783          & 0.229          & 21.18          & 0.698          & 0.233          \\
                             & 15                    & 25.43          & 0.768          & 0.242          & 20.99          & 0.650          & 0.249          \\ \hline
\end{tabular}
}
	\label{hyper weights}
\end{table*}

\subsubsection{Adaptive Sampling}
The comparison between rows 1 and 8 in Table~\ref{ablation} reveals that the adaptive sampling for density improves the PSNR by $1.66$ and $2.07$. 
It effectively captures valuable geometric information from sparse viewpoints, thereby generating fine-grained geometric details. 
The visualized ablation in Fig.~\ref{viz_ab} also demonstrates that when the baseline incorporates adaptive sampling outperforms in reconstructing objects distant from the camera (\textcolor{black}{black box}). This highlights its ability to resample complex objects and sparse boundaries to enhance rendering quality. 
Note that adaptive sampling can progressively correct edge errors during training, accelerating scene reconstruction and enhancing quality.

\subsection{Discussions}
 \subsubsection{Depth Model}
We adopt DPT~\cite{ranftl2021visiondpt}, a depth estimation model widely employed by state-of-the-art NeRF-based~\cite{wang2023sparsenerf} and GS-based works~\cite{zhu2023fsgs,li2024dngaussian}, to generate the initial coarse depth. To further evaluate the robustness of our proposed refined depth, we also performed ablation experiments using various DPT models, including \emph{dpt-hybrid-384} and \emph{dpt-large-384}, as shown in Table~\ref{depth_model}.

Our approach consistently achieves robust reconstruction results across different DPT variants, confirming the effectiveness of the refined depth strategy. In contrast, the state-of-the-art GS-based method FSGS~\cite{zhu2023fsgs} exhibits considerable variability in reconstruction performance depending on the coarse depth estimated by different DPT models. This finding indicates that employing more advanced or larger-scale depth estimation techniques does not necessarily improve reconstruction quality. Instead, by refining coarse depth into one characterized by accurate geometry and high-frequency details, our model effectively regularizes the image reconstruction process and demonstrates strong robustness.

\subsubsection{Initialization}
As depicted in Table~\ref{initialization} and Fig.~\ref{Comparision of the initialization}, we conducted visual ablation studies targeting different initialization methods, including the dense initialization based on COLMAP, Gaussian Unpooling initialization utilized by FSGS~\cite{zhu2023fsgs}, and our proposed adaptive sampling method. The results demonstrate a clear advantage of our proposed method of effective densification.
As shown in Fig.~\ref{Comparision of the initialization}, our proposed adaptive sampling initialization method can effectively address challenging areas such as boundaries of the original camera and regions difficult to render with Gaussian splats, achieving excellent results even for single backgrounds. In contrast, other initialization methods exhibit artifacts and blurry regions, resulting in black distortion effects, particularly noticeable in backgrounds with single colors like walls.

\subsubsection{Parameter of Spherical Harmonic (SH)}

For 3D Gaussian Splatting~\cite{kerbl20233dggs}, different spherical harmonics (SH) can yield varying colors in the reconstruction results. We conducted ablation experiments with different spherical harmonics on the NeRF-LLFF~\cite{mildenhall2019localllff} of 3 training views, as shown in Table~\ref{spherical harmonic}. The results from Table~\ref{spherical harmonic} demonstrate the strong robustness of our model to different spherical harmonics, indicating that our proposed refined depth and relative depth guidance methods enable the model to acquire strong geometric and detailed information accurately. Moreover, compared to the specially designed neural color renderer in DNGaussian~\cite{li2024dngaussian}, our model achieves superior reconstruction results.

\subsubsection{Hyperparameter of Energy Weights}

As shown in Table~\ref{hyper weights}, we performed an extensive experiment to investigate the impact of varying module weights $w_g$ and $w_h$  in Eqn.~\ref{energy} on the reconstruction quality. The best results were achieved when the weight $w_g$ was set to 10 and the weight $w_h$ to 5. It is worth noting that, given the foundational role of local geometric feature similarity, the weight $w_u$ was set as the baseline with a value of 1. The weight  $w_g$ governs global structural consistency, underscoring its pivotal role in constraining the depth map, thus necessitating a relatively higher value $10$. Conversely, $w_u$ modulates the alignment of high-frequency features to enhance edge preservation and suppress noise. However, the excessively high values $10$ and $15$ may overly prioritize local geometric similarity, potentially undermining the overall fidelity of geometric rendering.

\section{Conclusion and Future Work}
In this paper, we introduce RDG-GS, a real-time sparse-view 3D reconstruction method that utilizes relative depth
guidance by optimizing the spatial depth-image
similarity, thereby ensuring view-consistent geometry
reconstruction and fine-grained refinement.  It generates refined  depth priors and integrate global and local scene information
into Gaussians. It also employs adaptive sampling for quick and effective densification. RDG-GS achieves SOTA results in rendering quality and speed across four datasets. Codes will be released. 

\noindent \textbf{Limitations.} Although RDG-GS achieved excellent performance in sparse-view 3D rendering, we also considered the potential limitations, including: 1. 
While focusing on improving the quality of the reconstruction, it is also crucial to consider further optimizing the efficiency of the training. 2. The anisotropic shape of Gaussians complicates color and depth constraints in plane regions from sparse views, which may cause object artifacts. 3. While RDG-GS effectively learns scene geometry and fine details, further designs are needed to better handle mirror reflections.

\noindent\textbf{Future Work.} In future exploration, we plan to design an asymmetric approach employing multi-dimensional hierarchical Gaussians to achieve more efficient shape-structure rendering, while further optimizing depth-feature extraction to boost training efficiency and minimize overhead.
Additionally,
we will explore more sophisticated reflection models or incorporate domain-specific reflection priors to capture complex mirror-like surfaces. Such enhancements would improve the realism and accuracy of the reconstructed scenes, especially in environments with high-specular or reflective objects.
\section*{Data Availability Statements}

The datasets that support the findings of
this study are available in the following repositories: 
Mip-NeRF~\cite{barron2022mip260} dataset at \url{https://jonbarron.info/mipnerf360/}, NeRF-LLFF~\cite{mildenhall2019localllff} dataset at \url{https://bmild.github.io/llff/}, DTU~\cite{6909453DTU} dataset at \url{http://roboimagedata.compute.dtu.dk/?page_id=36}, and Blender dataset~\cite{mildenhall2021nerf} at \url{https://www.matthewtancik.com/nerf}.

\section*{Statements and Declarations}
\subsection*{Funding}
This work was supported in part by the Zhejiang Provincial
Natural Science Foundation of China under Grant LDT23F02023F02.

\bibliography{ref}
\bibliographystyle{abbrv}


\clearpage

\section*{Supplemental}
\subsection*{Detail Theoretical Analysis}
Our refined depth restoration hinges on utilizing high-quality RGB data to guide coarse depth, gradually aligning local structures and details near each pixel with the view-consistent geometry and fine details of the RGB image.
Thus, we compute the structural similarity of local patches centered at $i$ pixel between the depth map ${\boldsymbol{D}_r} = \{ {x_1},...,{x_n}\}$ and the RGB image $\boldsymbol{I} = \{ {y_1},...,{y_n}\}$. We first compute the local structure similarity through the local patches of the depth and the RGB image.
The local structure similarity module is defined based on the local structure similarity $SSIM(W_{i}^x, W_{i}^y)$ to encourage pixels in the refined depth map to be more likely to choose accurate and local fine-grained values:
\begin{equation}
\resizebox{0.85\hsize}{!}{$ {\psi _u}({r_i}) = \left\{ \begin{array}{l}
 - \log (SSIM(W_i^x,W_i^y)) \quad {r_i} = {x_i}\\
 - \log (\frac{1}{{l - 1}}(1 - SSIM(W_i^x,W_i^y)))\quad {r_i} \ne {x_i}
\end{array} \right.$}
\end{equation}
where $l$ represents the total count of potential intensities within the depth map.
For the global context information, we adopt the global structural consistency module to explore the correlation between the depth $\boldsymbol{D}_r$ and the RGB feature $\boldsymbol{I}$.  Neighboring pixels in images with similar colors suggest that co-located pixels in the depth map may have similar intensities. Thus, we devise the global structural consistency module as follows:
\begin{equation}
\resizebox{0.9\hsize}{!}{$ \psi_p(r_i,r_j)=(1-\exp (-\frac{ \mid x_i-x_j \mid ^2}{2 \theta_\mu^2})) \times \exp \left(-\frac{\|i-j \mid\|^2}{2 \theta_\alpha^2}-\frac{\|y_i-y_j\|^2}{2 \theta_\beta^2}\right)$}
\end{equation}

where $\theta_\alpha$ is the standard deviation of the global Gaussian kernel, $\theta_\mu$ and $\theta_\beta$ are for the local Gaussian kernel.
To refine the depth with correct geometric shapes and fine textures, we propose coarse and fine-grained models, each comprising $\psi _u$ and $\psi _p$ modules. The only distinction lies in the fine-grained model, where the $\theta _\alpha$, $\theta _\mu$, and $\theta _\beta$ are fixed to a smaller value, effectively mitigating texture replication artifacts generated by the coarse module.

\subsection*{Introduction of Datasets}
\label{sec:datasets}

\subsubsection*{Mip-NeRF360}
The Mip-NeRF360~\cite{barron2022mip260} dataset, comprises nine scenes, each showcasing a sophisticated central subject or locale set against an intricate backdrop. Following FSGS~\cite{zhu2023fsgs}, we specifically utilize the publicly available 8 scenes, including ``counter", ``room", ``bear", ``garden", ``bonsai", ``kitchen", ``bicycle", ``stump", and employing 24 training views with images downscaled to 2×, 4×, and 8× for comparison and others for testing. Test images are selected in accordance with the same protocol used for the NeRF-LLFF~\cite{mildenhall2019localllff} datasets. To the best of our knowledge, we are pioneering the exploration of novel view synthesis within unbounded scenes in Mip-NeRF360~\cite{barron2022mip260} with sparse-view inputs. 

\subsubsection*{NeRF-LLFF}
The LLFF dataset~\cite{mildenhall2019localllff} comprises 8 real-world scenes facing forward, including ``fern", ``flower", ``fortress", ``horns", ``leaves", ``orchids", ``room", ``trex". Following the approach of RegNeRF~\cite{niemeyer2022regnerf} and FSGS~\cite{zhu2023fsgs}, we select one image from every 8 images as the test set and evenly sample sparse views from the remaining images for training. 
We utilize a training setup with 3 views and assess performance with 8 views, considering resolutions of both 1008 $\times$ 756 and 504 $\times$ 378.

\subsubsection*{DTU}
The DTU dataset~\cite{6909453DTU} comprises 124 scenes focused on individual objects, captured through a fixed camera setup. We adhere to the methodology outlined in RegNeRF~\cite{niemeyer2022regnerf} and SparseNeRF~\cite{wang2023sparsenerf} to assess models directly across 15 specific scenes, identified by scan IDs 8, 21, 30, 31, 34, 38, 40, 41, 45, 55, 63, 82, 103, 110, and 114. Within each scan, images assigned the IDs 25, 22, and 28 serve as the input views within our 3-view configuration. For evaluation purposes, the test set encompasses images with IDs 1, 2, 9, 10, 11, 12, 14, 15, 23, 24, 26, 27, 29, 30, 31, 32, 33, 34, 35, 41, 42, 43, 45, 46, and 47, all of which undergo a 4× downsampling.

\subsubsection*{Blender}
For the Blender dataset~\cite{mildenhall2021nerf}, we adhere to the data segmentation approach utilized in Freenerf~\cite{yang2023freenerf}. We selected 8 training input views identified by the IDs 26, 86, 2, 55, 75, 93, 16, 73 and  25 test views for evaluation. Throughout the experimentation, all images are downsampled by a factor of 2 to dimensions of 400 × 400.

\end{document}